\documentclass[final]{cvpr}

\usepackage{times}
\usepackage{epsfig}
\usepackage{graphicx}
\usepackage{amsmath}
\usepackage{amssymb}

\usepackage{graphicx}
\usepackage{comment}
\usepackage{amsmath,amssymb} 
\usepackage{color}
\usepackage[dvipsnames]{xcolor}
\usepackage{multirow}

\usepackage{caption}
\usepackage{subcaption}
\usepackage{booktabs}
\usepackage{tabularx}

\newcommand{\KG}[1]{{\color{black}#1}}

\usepackage[pagebackref=true,breaklinks=true,colorlinks,bookmarks=false]{hyperref}

\def\cvprPaperID{5741} 
\def\confYear{CVPR 2021}

\begin{document}

\title{Fashion IQ: A New Dataset Towards \\ Retrieving Images by Natural
Language Feedback}


\author{
Hui Wu\thanks{\,\,Equal contribution.}$^{\,\,1,2}$ \quad Yupeng Gao$^{*2}$ \quad Xiaoxiao Guo$^{*2}$ \quad Ziad Al-Halah$^{3}$\vspace{0.05in} \\ 
Steven Rennie$^{4}$ \quad  Kristen Grauman$^{3}$ \quad Rogerio Feris$^{1,2}$
\\ \\
   $^1$ MIT-IBM Watson AI Lab \quad
   $^2$ IBM Research \quad 
   $^3$ UT Austin \quad
   $^4$ Pryon 
   \\
  }

\maketitle

\begin{abstract}
Conversational interfaces for the detail-oriented retail fashion domain are more natural, expressive, and user friendly than classical keyword-based search interfaces. 
In this paper, we introduce the Fashion IQ dataset to support and advance research on interactive fashion image retrieval.
Fashion IQ is the first fashion dataset to provide human-generated captions that distinguish similar pairs of garment images  together with side-information consisting of real-world product descriptions and derived visual attribute labels for these images. 
We provide a detailed analysis of the characteristics of the Fashion IQ data, and present a transformer-based user simulator and interactive image retriever that can seamlessly integrate visual attributes with image features, user feedback, and dialog history, leading to improved performance over the state of the art in dialog-based image retrieval. We believe that our dataset will encourage further work on developing more natural and real-world applicable conversational shopping assistants.\footnote{Fashion IQ is available at https://github.com/XiaoxiaoGuo/fashion-iq}

\end{abstract}

\begin{figure}[th!]
\begin{center}
\includegraphics[width=\linewidth]{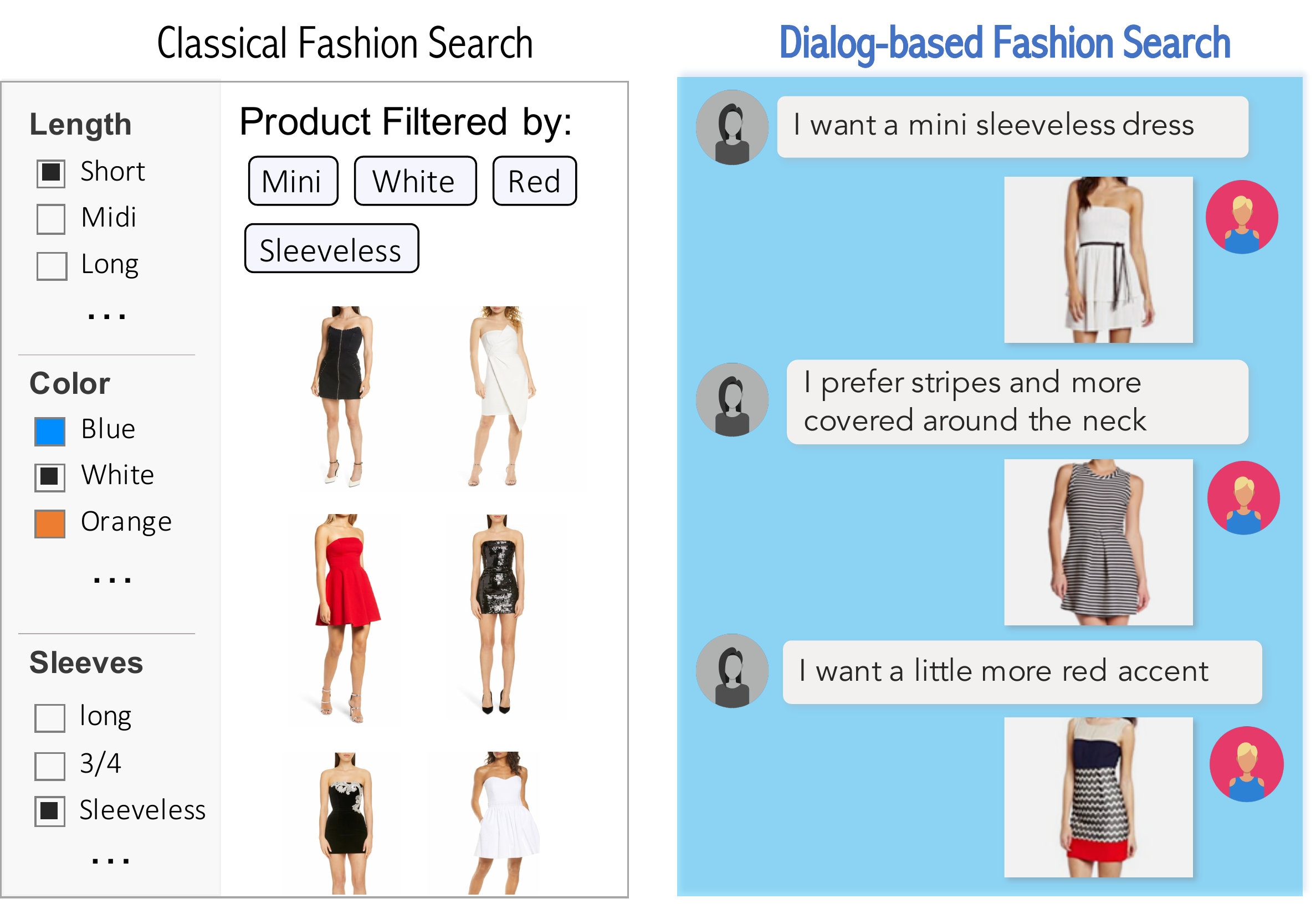}

\end{center}
\vspace{-1.5em}
\caption{
    A classical fashion search interface relies on the user selecting filters based on a pre-defined fashion ontology.
    This process can be cumbersome and the search results still need manual refinement.
    The Fashion IQ dataset supports building dialog-based fashion search systems, which are more natural to use and allow the user to precisely describe what they want to search for.
    }
\label{fig:teaser}
\vspace{-1.5em}
\end{figure}

\vspace{-1em}
\section{Introduction}
Fashion is a multi-billion-dollar industry, with direct social, cultural, and economic implications in the world. Recently, computer vision has demonstrated remarkable success in many applications in this domain, including trend forecasting \cite{al2017fashion}, creation of capsule wardrobes \cite{hsiao2018creating}, interactive product retrieval \cite{guo2018dialog,yang2017visual}, recommendation \cite{mcauley2015image}, and fashion design \cite{rostamzadeh2018fashion}.

In this work, we address the problem of interactive image retrieval for fashion product search.
High fidelity interactive image retrieval, despite decades of research and many great strides, remains a research challenge. At the crux of the challenge are two entangled elements: empowering the user with ways to express what they want, and  empowering the retrieval machine with the information, capacity, and learning objective to realize high performance.

To tackle these challenges, traditional systems have relied on relevance feedback \cite{rui1998relevance,yang2017visual},
allowing users to indicate which images are ``similar'' or ``dissimilar'' to the desired image. Relative attribute feedback (e.g., ``more formal than these'', ``shinier than these'') \cite{kovashka2012,kovashka2017attributes} allows the comparison of the desired image with candidate images based on a fixed set of attributes. While effective,  this specific form of user feedback constrains what the user can convey. 

Recent work on image retrieval has demonstrated the power of utilizing natural language to address this problem~\cite{vo2018composing,guo2018dialog,tan2019},
with relative captions describing the differences between a reference image and what the user has in mind, and dialog-based interactive retrieval as a principled and general methodology for interactively engaging the user in a multimodal \emph{conversation} to resolve their intent \cite{guo2018dialog}.
When empowered with natural language feedback, the user is not bound to a pre-defined set of attributes, and can communicate compound and more specific details during each query, which leads to more effective retrieval.
For example, with the common attribute-based interface (Figure~\ref{fig:teaser} left) the user can only define what kind of attributes the garment has (e.g., white, sleeveless, mini), however with interactive and relative natural language feedback (Figure~\ref{fig:teaser} right) the user can use comparative forms (e.g., more covered, brighter) and fine-grained compound attribute descriptions (e.g., red accent at the bottom, narrower at the hips).

While this recent work represents great progress, several important questions remain.
In real-world fashion product catalogs, images are often associated with \emph{side information}, which in the wild varies greatly in format and information content, and can often be acquired at large scale with low cost. 
Furthermore, often descriptive
representations such as attributes can be extracted from this data, and form a strong basis for generating stronger image captions \cite{you2016image,wu2018image,yao2017boosting} and more effective image retrieval \cite{huang2015cross,berg2010automatic,simo2016fashion,laenen2017cross}.
How such side information interacts with natural language user inputs, and how it can be best used to improve the state of the art dialog-based image retrieval systems are important open research questions. 

State-of-the-art conversational systems currently typically require cumbersome hand-engineering and/or large-scale dialog data~\cite{li2018towards,budzianowski2018}.
In this paper, we investigate the extent to which side information can alleviate these requirements, and incorporate side information in the form of visual attributes into model training to realize  improved user simulation and interactive image retrieval. This represents an important step toward the ultimate goal of \KG{constructing}
commercial-grade conversational interfaces  with much less data and effort, and much wider real-world applicability. 

Toward this end, we contribute a new dataset, Fashion Interactive Queries (\emph{Fashion IQ}) and explore methods for jointly leveraging natural language feedback and side information to realize effective and practical image retrieval systems (see Figure \ref{fig:teaser}). 
Fashion IQ is situated in the detail-critical fashion domain, where expressive conversational interfaces have the potential to dramatically improve the user experience.  
Our main contributions are as follows:

\begin{itemize}
\vspace{-0.5em}
\item We introduce a novel dataset, Fashion IQ, which we will make publicly available as a new resource for advancing research on conversational fashion retrieval. 
Fashion IQ is the first fashion dataset that includes both human-written relative captions that have been annotated for similar pairs of images, and the associated real-world product descriptions and attribute labels for these images as side information.  

\vspace{-0.5em}
\item We present \KG{a} transformer-based user simulator and interactive image retriever that can seamlessly leverage multimodal inputs (images, natural language feedback, and attributes) during training, and \KG{leads} to significantly improved performance. Through the use of self-attention, these models consolidate the traditional components of user modeling and interactive retrieval, are highly extensible, and outperform existing methods for the relative captioning and interactive image retrieval of fashion images on Fashion IQ.

\vspace{-0.5em}
\item To the best of our knowledge, this is the first study \KG{to} investigate the benefit of combining natural language user feedback and attributes for dialog-based image retrieval, and \KG{it} provides empirical evidence that incorporating attributes results in superior performance for both user modeling and dialog-based image retrieval.
\vspace{-0.5em}
\end{itemize}
\section{Related Work }

\textbf{Fashion Datasets.} Many fashion datasets have been proposed over the past few years, covering different applications such as fashionability and style prediction \cite{simo2015neuroaesthetics,kiapour2014hipster,hsiao2017learning,simo2016fashion}, fashion image generation \cite{rostamzadeh2018fashion}, product search and recommendation \cite{huang2015cross,yu2014fine,hadi2015buy,mcauley2015image,veit2015learning}, fashion apparel pixelwise segmentation \cite{jia2020fashionpedia,zheng2018modanet,yang2014clothing}, and body-diverse clothing recommendation \cite{hsiao2020vibe}. DeepFashion \cite{DeepFashion,ge2019deepfashion2} is a large-scale fashion dataset containing consumer-commercial image pairs and labels such as clothing attributes, landmarks, and segmentation masks.  
iMaterialist \cite{guo2019imaterialist} is a large-scale dataset with fine-grained clothing attribute annotations, while Fashionpedia \cite{jia2020fashionpedia} has both attribute labels and corresponding pixelwise segmented regions.  

Unlike most existing fashion datasets used for image retrieval, which focus on content-based or attribute-based product search, our proposed dataset facilitates research on
{\em conversational} fashion image retrieval. In addition,
we 
\KG{enlist}
real users to collect the high-quality, {\em }natural language annotations, rather than using fully
or partially automated approaches to acquire large amounts of
weak attribute labels~\cite{mcauley2015image,DeepFashion,rostamzadeh2018fashion} or synthetic conversational data~\cite{saha2018towards}.
Such high-quality annotations are more costly, but of great benefit in building and evaluating conversational systems for image retrieval. We make the data publicly available so that the community can explore
the value of combining high-quality human-written relative captions and the more common, web-mined
weak annotations.

\vspace{0.1em}
\textbf{Visual Attributes for Interactive Fashion Search.} 
Visual attributes, including color, shape, and texture, have been successfully used to model clothing images \cite{huang2015cross,hsiao2017learning,hsiao2018creating,al2017fashion,zhao2017memory,chen2015deep,lu2017fully}. 
More relevant to our work, in \cite{zhao2017memory}, a system for interactive fashion search with attribute manipulation was presented, where the user can choose to modify a query by changing the value of a specific attribute. 
While visual attributes model the presence of certain visual properties in images,  they do not measure the relative strength of them. 
To address the issue, relative attributes \cite{parikh2011relative,souri2016deep} were proposed, and have been exploited as a richer form of feedback for interactive fashion image retrieval \cite{kovashka2017attributes,kovashka2012,kovashka2013attribute,shades2014}. 
However, in general, attribute based retrieval interfaces require careful curation and engineering of the attribute vocabulary. 
Also, 
when attributes are used as the sole interface for user queries, they can lead to inferior performance relative to both relevance feedback~\cite{plummer2019give} and  natural language feedback~\cite{guo2018dialog}.
In contrast with attribute based systems, our work explores the use of relative feedback in \emph{natural language}, which is more flexible and expressive, and is complementary to attribute based interfaces.

\begin{figure}
\begin{center}
\includegraphics[width=\linewidth]{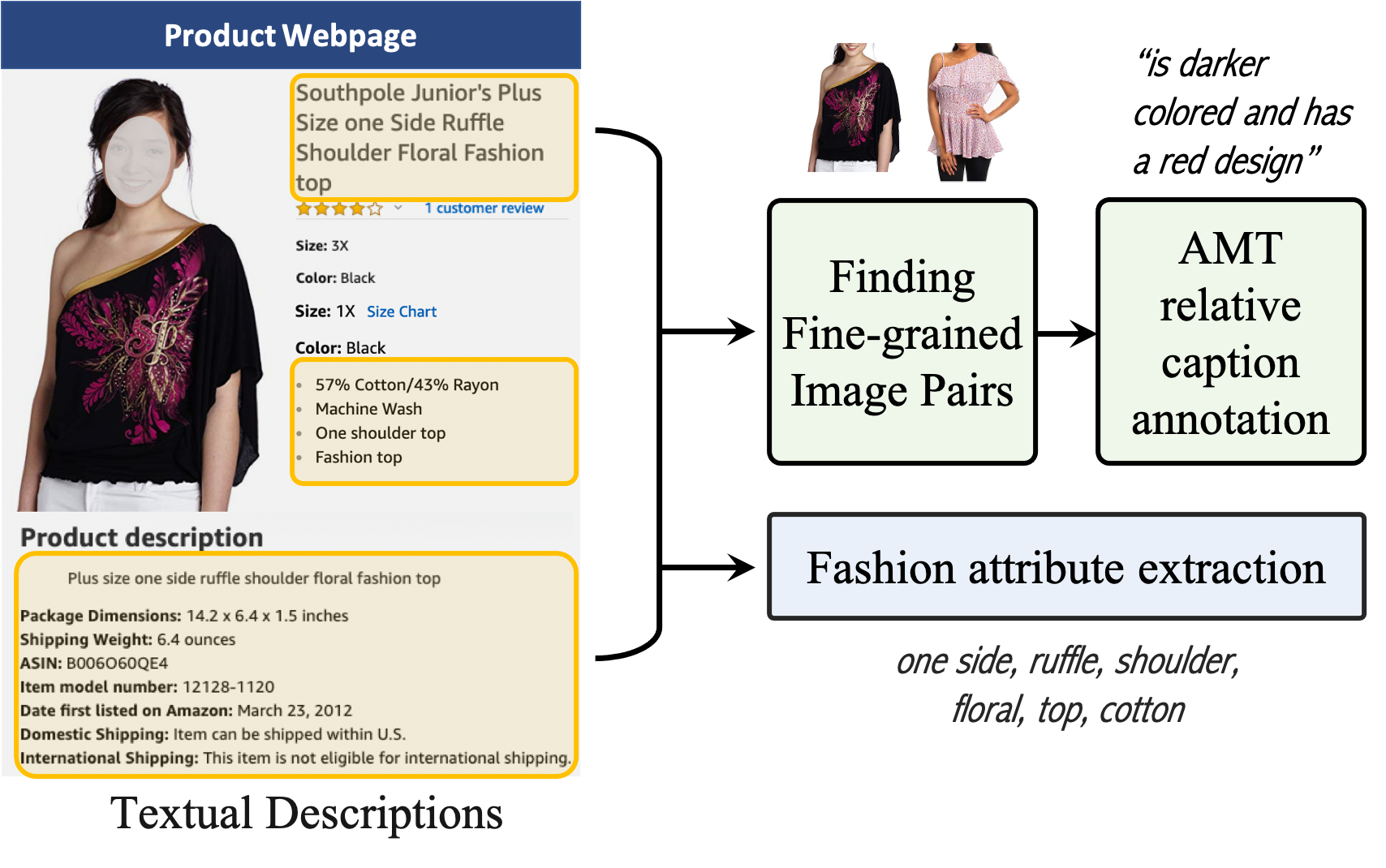}
\end{center}
\vspace{-2em}
   \caption{Overview of the dataset collection process.}
    \label{fig:collection_process}
    \vspace{-2em}
\end{figure}

\vspace{0.1em}
\textbf{Image Retrieval with Natural Language Queries.}
Methods that lie in the intersection of computer vision and natural language processing, including image captioning \cite{rennie2016self,vinyals2015show,icml2015_xuc15} and visual question-answering \cite{VQA,das2018embodied,tapaswi2016movieqa}, have received much attention from the research community. Recently, several techniques have been proposed for image or video retrieval based on natural language queries \cite{li2017person,barbu2013saying,tellex2009towards,vo2018composing,tan2019}. 
In another line of work, visually-grounded dialog systems \cite{das2016visual,strub2017end,de2016guesswhat,das2017learning} have been developed to hold a meaningful dialog with humans in natural, conversational language about visual content. Most current systems, however, are based on purely text-based questions and answers regarding a single image. 
Similar to \cite{guo2018dialog}, we consider the setting of {goal-driven} dialog, where the user provides feedback in natural language, and the agent outputs retrieved images. 
Unlike \cite{guo2018dialog}, we provide a large dataset of relative captions anchored with real-world contextual information, which is made available to the community. 
In addition, we follow a very different methodology based on a unified  transformer model, instead of fragmented components to model the state and flow of the conversation, and show that the joint modeling of visual attributes and relative feedback via natural language can improve the performance of interactive image retrieval.

\vspace{0.05in}
\textbf{Learning with Side Information.} Learning with privileged information that is available at training time but not at test time is a popular machine learning paradigm \cite{vapnik2009new}, with many applications in computer vision \cite{sharmanska2013learning,huang2015cross}. In the context of fashion, \cite{huang2015cross} showed that visual attributes mined from online shopping stores serve as useful privileged information for cross-domain image retrieval. Text surrounding fashion images has also been used as side information to discover attributes \cite{berg2010automatic,han2017automatic}, learn weakly supervised clothing representations \cite{simo2016fashion}, and improve search based on noisy and incomplete product descriptions \cite{laenen2017cross}. In our work, for the first time, we explore the use of side information in the form of visual attributes for image retrieval with a natural language feedback interface.

\begin{figure*}
     \centering
     \begin{subfigure}[b]{\textwidth}
         \centering
         \includegraphics[width=0.85\textwidth]{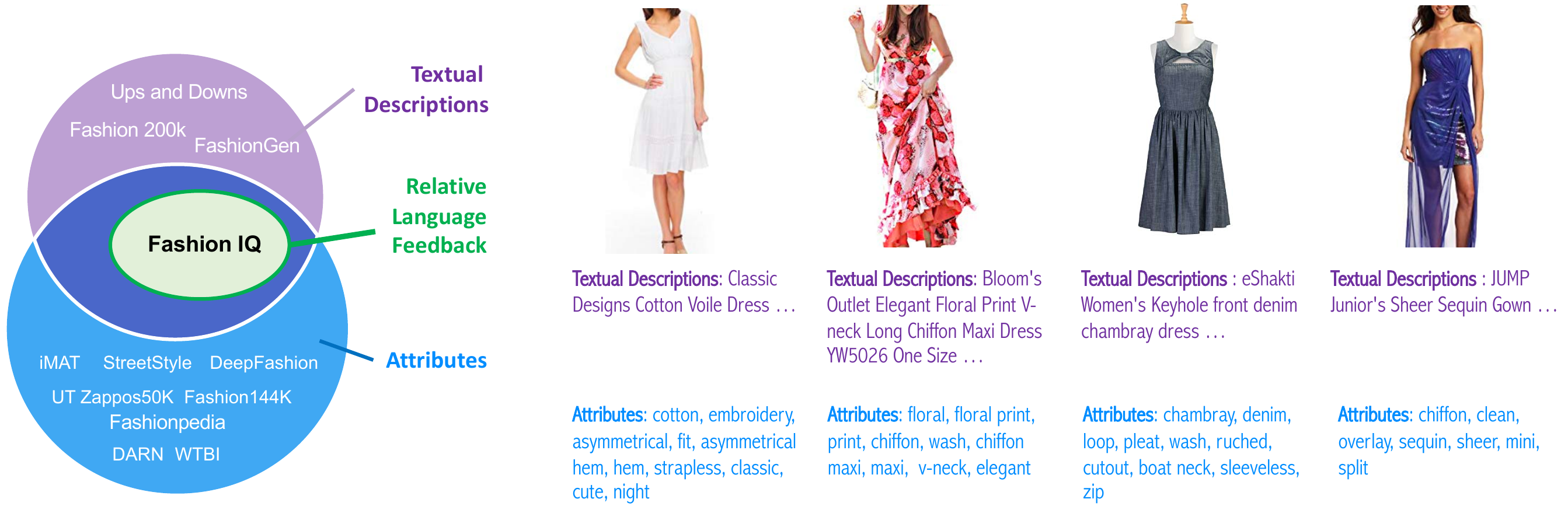}
         \caption{}
         \label{fig:dataset_overviewa}
         \vspace{-0.3em}
     \end{subfigure}
     \hfill
     \begin{subfigure}[b]{\textwidth}
         \centering
         \includegraphics[width=0.85\textwidth]{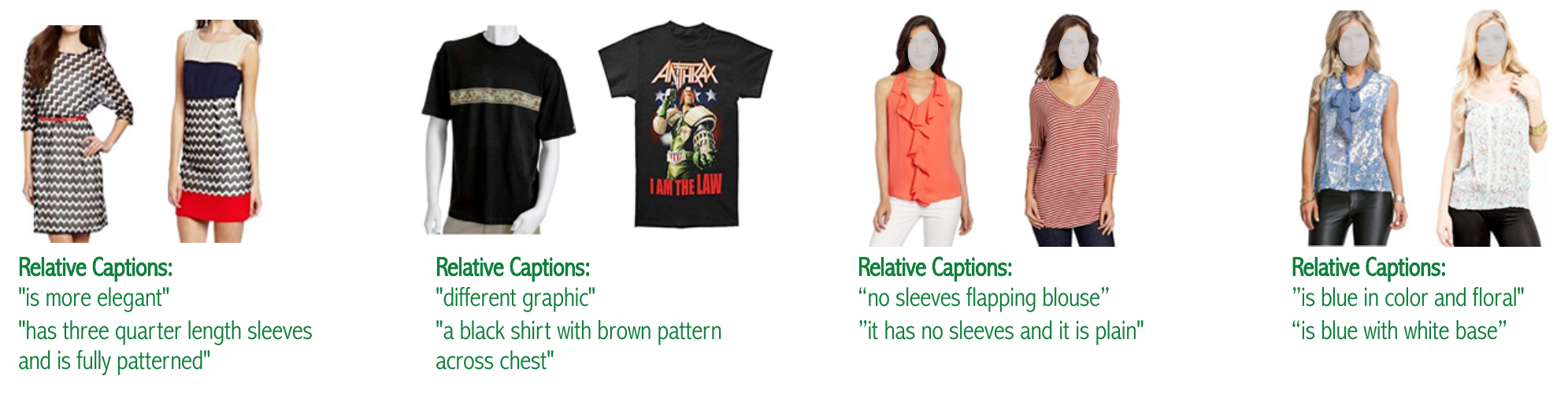}
         \caption{}
         \label{fig:dataset_overviewb}
         \vspace{-0.5em}
     \end{subfigure}
     \vspace{-1.5em}
\caption{Our Fashion IQ dataset is uniquely positioned to provide a valuable resource for research in joint modeling of user relative feedback via natural language and fashion attributes to develop interactive dialog-based retrieval models. (a) examples of the textual descriptions and attribute labels; (b) examples of relative captions. 
}
	\label{fig:dataset_overview}
	\vspace{-1em}
\end{figure*}

\section{Fashion IQ Dataset}

One of our main objectives in this work is to provide researchers with a strong resource for developing interactive dialog-based fashion retrieval models.
To that end, we introduce a novel public benchmark, Fashion IQ.
The dataset contains diverse fashion images (dresses, shirts, and tops\&tees), side information in form of textual descriptions and product meta-data, attribute labels, and most  importantly, large-scale annotations of high quality relative captions collected from human annotators.
Next we describe the data collection process and provide an in-depth analysis of Fashion IQ. The overall data collection procedure is illustrated in Figure~\ref{fig:collection_process}. Basic statistics of the resulting Fashion IQ dataset are summarized in
Table~\ref{table:statistics}.

\begin{table}
\begin{center}
\scalebox{0.8}{
\begin{tabular}{lccc}
\toprule
     \,\,\,\,\,& \textbf{\#Image} & \textbf{\# With Attr.} & \textbf{\# Relative Cap.} \\
     \midrule
     \multicolumn{4}{l}{\textbf{Dresses}}\\
     \multicolumn{1}{p{1.6cm}}{Train} & 11,452 & 7,741 & 11,970 \\
     Val & 3,817 & 2,561 & 4,034 \\
     Test & 3,818 & 2,653 & 4,048 \\
     Total & 19,087 & 12,955 & 20,052\\
     \midrule
     \multicolumn{4}{l}{\textbf{Shirts}}\\
     Train & 19,036 & 12,062 & 11,976 \\
     Val & 6,346 & 4,014 & 4,076 \\
     Test & 6,346 & 3,995 & 4,078 \\
     Total & 31,728 & 20,071 & 20,130 \\
     \midrule
\multicolumn{4}{l}{\textbf{Tops\&Tees}}\\
Train & 16,121 & 9,925 & 12,054 \\
Val & 5,374 & 3,303 & 3,924 \\
Test & 5,374 & 3,210 & 4,112 \\
Total & 26,869 & 16,438 & 20,090\\
     \bottomrule
\end{tabular}
}
\vspace{-1em}
\caption{Dataset statistics on Fashion IQ.\label{table:statistics}}
\end{center}
\vspace{-2em}
\end{table}

\subsection{Image And Attribute Collection}
The images of fashion products that comprise Fashion IQ were originally sourced from a product review dataset~\cite{he2016ups}.
Similar to~\cite{al2017fashion}, we selected three categories of product items, 
specifically: Dresses, Tops\&Tees, and Shirts.
For each image, we followed the link to the product website available in the dataset, in order to extract corresponding product information, when available.

\KG{Leveraging} the rich
textual information contained in the product website, \KG{we} extracted fashion attribute labels from them. More specifically, product attributes were extracted from the product title, the product summary, and detailed product description.
To define the set of product attributes, we adopted the fashion attribute vocabulary curated in DeepFashion~\cite{DeepFashion}, which is currently the most widely adopted benchmark for fashion attribute prediction. In total, this resulted in 1000 attribute labels, which were further grouped into five attribute types: texture, fabric, shape, part, and style. We followed a similar procedure as in \cite{DeepFashion} to extract the
attribute labels: an attribute label for an image is considered as present if its associated
attribute word appears at least once in the metadata. In Figure~\ref{fig:dataset_overviewa}, we provide examples of the 
original side information obtained from the product reviews and the corresponding attribute labels that were extracted.
To complete and denoise attributes, we use an attribute prediction model pretrained on DeepFashion attributes. The details are in Appendix~\ref{app:attribute_prediction}.

\subsection{Relative Captions Collection}
The Fashion IQ dataset is constructed with the goal of advancing conversational image
search. Imagine a typical visual search process (illustrated in Figure~\ref{fig:teaser}): a
user might start the search by describing general keywords which can weed out
totally irrelevant search instances, then the user can construct natural language
phrases which are powerful in specifying the subtle differences between the search target and the current search result. In other words, relative
captions are more effective to narrow down fine-grained cases than using
keywords or attribute label filtering.

To ensure that the relative captions can describe the fine-grained visual differences between
the reference and target image, we leveraged  product title information to select similar images for annotation with relative captions. Specifically, we first computed the TF-IDF score of all words appearing in each product title, and then for each target image, we paired it with a reference image by finding the image in the database (within the same data split subset) with the maximum sum of the TF-IDF weights on each overlapping word.
We randomly selected $\sim$10,000 target images
for each of the three fashion categories, and collected two sets of captions for each pair. Inconsistent captions were filtered (please consult the suppl. material for details). 

To amass relative captions for the Fashion IQ data, we  collected data using crowdsourcing. Briefly, the users were situated in the context of an online shopping chat window, and assigned the goal of providing a natural language expression to communicate to the shopping assistant the visual features of the search target as compared to the provided search candidate. Figure~\ref{fig:dataset_overviewb} shows examples of image pairs presented to the user, and the
resulting relative image captions that were collected.
We only included workers from three predominantly  English-speaking countries, with master level of expertise as defined in the crowdsourcing tool
and with an acceptance rate above 95\%. This criterion makes it costly to obtain the captions, but
ensures that the human-written captions in Fashion IQ are indeed of high quality.
To further improve the quality of the annotations and speed up the annotation process, the prefix of the relative feedback ``Unlike the provided image, the one I want" \KG{is} provided with the prompt, and the user only needs to provide a phrase \KG{that} focuses on the visual differences of the given image pairs.

\subsection{Dataset Analysis}
\begin{figure}
	\centering
	\includegraphics[width=\linewidth]{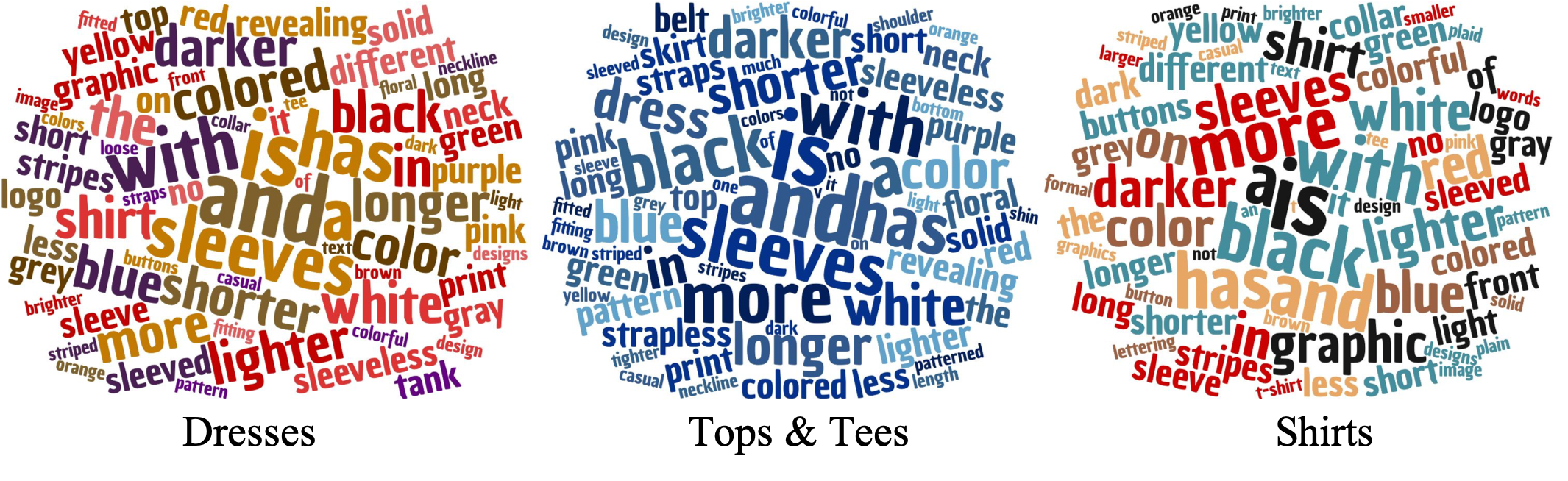}
	\vspace{-2em}
	\caption{Vocabulary of relative captions scaled by frequency}
	\vspace{-1em}
	\label{fig:cloud}
\end{figure}

Figure~\ref{fig:dataset_overviewb} depicts examples of collected relative captions in the Fashion IQ dataset, and Figure~\ref{fig:cloud} displays word-frequency clouds of the relative captions in each fashion category. The natural language based data annotation process results in rich fashion vocabularies for each subtask, with prominent visual differences often being implicitly agreed upon by both annotators, and resulting in semantically related descriptions. The empirical distributions of relative caption length and number of
attributes per image for all subsets of Fashion IQ are similarly distributed across
all three datasets.\footnote{c.f. Figure~\ref{fig:distribution} in the Appendix.} In most cases,
the attribute labels and relative captions contain complementary information, and thus jointly form a stronger basis for ascertaining the relationships between images.

\begin{table}
\begin{center}
\scalebox{0.88}{
\begin{tabular}{ccc}
\toprule
\textbf{Semantics} & \textbf{Quantity} & \textbf{Examples}\\
\midrule
Direct reference  &  49\%
& \multicolumn{1}{p{4.2cm}}{is solid white and buttons up with front pockets} \\
Comparison & 32\%
& \multicolumn{1}{p{4.2cm}}{\textbf{has longer sleeves} and is \textbf{lighter in color}} \\
Direct \& compar. & 19\%
& \multicolumn{1}{p{4.2cm}}{has a geometric print with \textbf{longer sleeves}} \\
\midrule
Single attribute & 30.5\%
& \multicolumn{1}{p{4.2cm}}{\textbf{is more bold}} \\
Composite attr. & 69.5\%
& \multicolumn{1}{p{4.2cm}}{black \textbf{with} red cherry pattern \textbf{and} a deep V neck line} \\
\midrule
Negation & 3.5\%
& \multicolumn{1}{p{4.2cm}}{is white colored with a graphic and \textbf{no lace design}} \\
\bottomrule
\end{tabular}
}
\end{center}
\vspace{-1.5em}
\caption{Analysis on the relative captions. Bold font highlights comparative phrases between the target and the reference images.}
\label{table:analysis}
\vspace{-1em}
\end{table}

\vspace{0.5em}
{\bf \noindent Comparing relative captions and attributes.}
To further obtain insight on the unique properties of the relative
captions in comparison with classical attribute labels, we conducted a semantic analysis on a subset of 200 randomly chosen relative
captions. The results of the analysis are summarized in Table~\ref{table:analysis}.
Almost 70\% of all text queries in Fashion IQ consist of
compositional attribute phrases. Many of
the captions are simpler adjective-noun pairs (e.g. “red
cherry pattern”). Nevertheless, this structure is more complex
than a simple ”bag of attributes” representation, which
can quickly become cumbersome to build, necessitating a large vocabulary and
compound attributes, or multi-step composition. Furthermore,
in excess of 10\% of the data involves more complicated
compositions that often include direct or relative
spatial references for constituent objects (e.g. “pink
stripes on side and bottom”).
The analysis suggests that relative captions are a more expressive and flexible form of
annotation than attribute labels, which are commonly provided in previous fashion datasets.
The diversity in the structure and content of the relative captions provide
a fertile resource for modeling user
feedback and for learning natural language feedback based image retrieval models, \KG{as we will demonstrate below.}

\subsection{Fashion IQ Applications\label{sec:applications}}

\begin{figure*}
\centering
\includegraphics[width=0.85\linewidth]{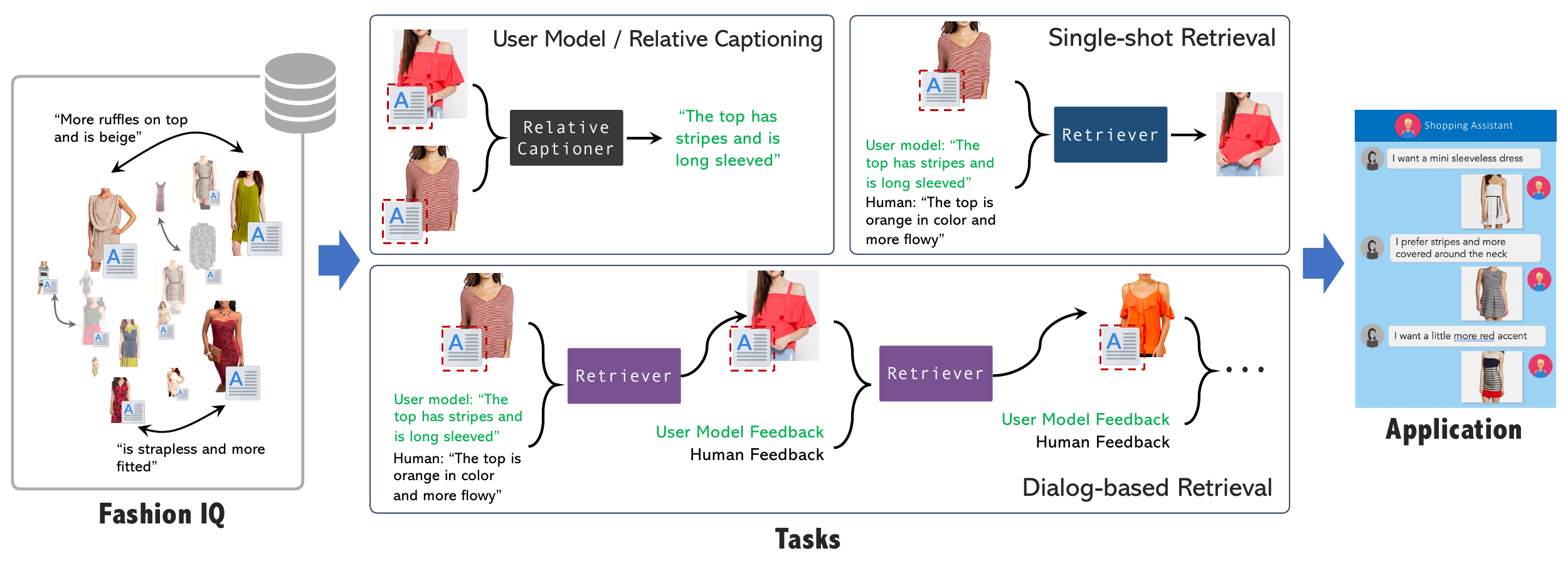}
\vspace{-1em}
\caption{Fashion IQ can be used in different scenarios to enhance the development of
an interactive fashion retrieval system with natural language interaction. We provide
three example scenarios: user modeling and two types of
retrieval tasks.
Fashion IQ uniquely provides both annotated user feedback (black font) and visual attributes derived from real-world product data (dashed boxes) for system training.}
\label{fig:tasks}
\vspace{-1em}
\end{figure*}

The Fashion IQ dataset can be used in different ways to drive progress on developing
more effective interfaces for image retrieval (as shown in Figure~\ref{fig:tasks}). 
These tasks can be developed as standalone applications, or can be investigated in conjunction. Next, we briefly introduce the component tasks associated with developing interactive image retrieval applications, and discuss how Fashion IQ can be utilized to realize and enhance these components. 

\emph{Single-shot Retrieval.}
Single-turn image retrieval systems have now evolved to support multimodal queries that include both images and text feedback. Recent work, for example, has attempted to use natural language feedback to modify a visual query~\cite{cheneccv2020,dodds2020modality,chen2020image}. By virtue of human-annotated 
relative feedback sentences, Fashion IQ serves as a rich resource for multimodal search
using natural language feedback. We provide \KG{an} additional study 
using Fashion IQ to train single-shot retrieval systems in Appendix~\ref{sec:retrieve}.

\emph{Relative Captioning.}
The relative captions of Fashion IQ \KG{make} it a valuable resource to train and evaluate relative
captioning systems~\cite{jhamtani2018learning,tan2019expressing,park2019robust,forbes2019neural}. In particular, when applied to conversational image search, a
relative captioner can be used as a user model to provide a large amount of low-cost training data for dialog models. 
Fashion IQ introduces the opportunity to utilize both attribute labels and human-annotated relative captions to train stronger user simulators, and correspondingly stronger interactive image retrieval systems.  In the next section, we introduce a strong baseline model for relative captioning and
demonstrate how it \KG{can} 
be leveraged as a user model to assist the training of a dialog-based interactive retriever. 

\emph{Dialog-based Interactive Image Retrieval.}
Recently, dialog-based interactive image retrieval~\cite{guo2018dialog} has been proposed as a new interface and framework for interactive image retrieval. 
Fashion IQ with the large scale data ($\sim$6x larger), the additional attribute labels, and the more diverse set of fashion categories, 
allows for more comprehensive research on interactive product retrieval systems. 
We will show next, how the different modalities available in Fashion IQ can be incorporated together effectively using a multimodal transformer to build a state-of-the-art dialog-based image retrieval model.

\vspace{-0.05in}
\section{Multimodal Transformers for Interactive Image Retrieval}
To advance research on the Fashion IQ applications, we introduce a strong baseline for dialog-based fashion retrieval based on the modern transformer architecture~\cite{vaswani2017attention}.
Multimodal transformers have recently received significant attention, achieving state-of-the-art results in vision and language tasks such as image captioning and visual-question answering \cite{zhou2020unified,tan2019lxmert,li2020oscar,sun2019videobert,lu2019vilbert}. 
To the best of our knowledge, multimodal transformers have not been studied in the context of goal-driven dialog-based image retrieval.
We adapt the transformer architecture in a multimodal framework that incorporates image features, fashion attributes, and a user's textual feedback in a unified approach.
Our model architecture allows for more flexibility in terms of included modalities compared to the RNN-based approaches (e.g.,~\cite{guo2018dialog}) 
which may require a systemic revision whenever a new modality is included. For example, integrating visual attributes into traditional goal-driven dialog architectures would require specialization of each individual component to model the user response, track the dialog history, and generate responses.
Next we describe our relative captioner transformer, which is then used as a user simulator to train our interactive retrieval system.

\vspace{-0.05in}
\subsection{Relative Captioning Transformer}
\begin{figure}[t!]
	\centering
	\includegraphics[width=0.85\linewidth]{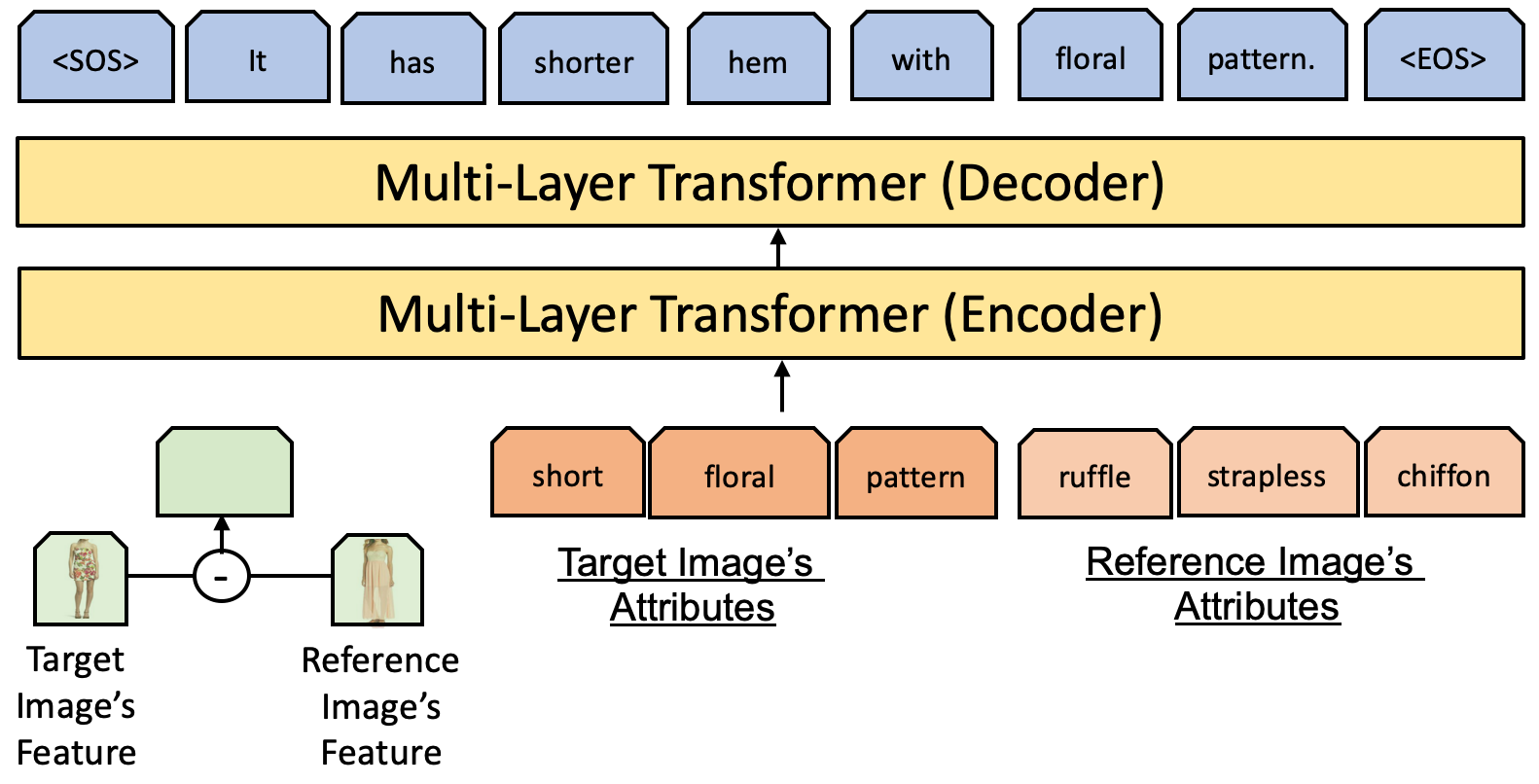}
	
	\caption{Our multimodal transformer model for relative captioning, which is used as a user simulator for training our interactive image retrieval system.\label{fig:tf_cap}}
	\vspace{-1em}
\end{figure}

As discussed earlier in Sec.~\ref{sec:applications}, in the relative captioning task the model is given a reference image $I_r$ and a target image $I_t$ and it is tasked with describing the differences of $I_r$ relative to $I_t$ in natural language.
Our transformer model leverages two modalities: image visual feature and \KG{inferred} attributes (Figure~\ref{fig:tf_cap}).
While the visual features capture the fine-grained differences between $I_r$ and $I_t$, the attributes help in highlighting the prominent  differences between the two garments.
Specifically, we encode each image with a CNN encoder $f_I(\cdot)$, and to obtain the prominent set of fashion attributes from each image, we use an attribute prediction model $f_A(\cdot)$ and select the top $N=8$ predicted attributes from the reference $\{a_i\}^r$ and the target $\{a_i\}^t$ images based on confidence scores from $f_A(I_r)$ and $f_A(I_t)$, respectively.
Then, each attribute is embedded into a feature vector based on the word encoder $f_W(\cdot)$.
Finally, our transformer model attends to the difference in image features of $I_r$ and $I_t$ and their attributes to produce the relative caption $\{w_i\} = f_R(I_r, I_t) = (f_I(I_r) - f_I(I_t), f_W(\{a_i\}^r), f_W(\{a_i\}^t))$, where $\{w_i\}$ is the word sequence generated for the caption.

\vspace{-0.05in}
\subsection{Dialog-based Image Retrieval Transformer}

In this interactive fashion retrieval task, the user provides textual feedback based on the currently retrieved image to guide the system towards a target image during each interaction turn (in the form of a relative caption describing the differences between the retrieved image and the image the user has in mind). 
At the end of each turn, the system then responds with a new retrieved image, based on all of the user feedback received so far.
Here we adopt a transformer architecture that enables our model to attend to the entire, multimodal history of the dialog during each dialog turn. This is in contrast with RNN-based models (e.g.,~\cite{guo2018dialog}), which must systemically incorporate features from different modalities, and consolidate historical information into a low-dimensional feature vector.

\begin{figure}[t]
	\centering
	\includegraphics[width=0.9\linewidth]{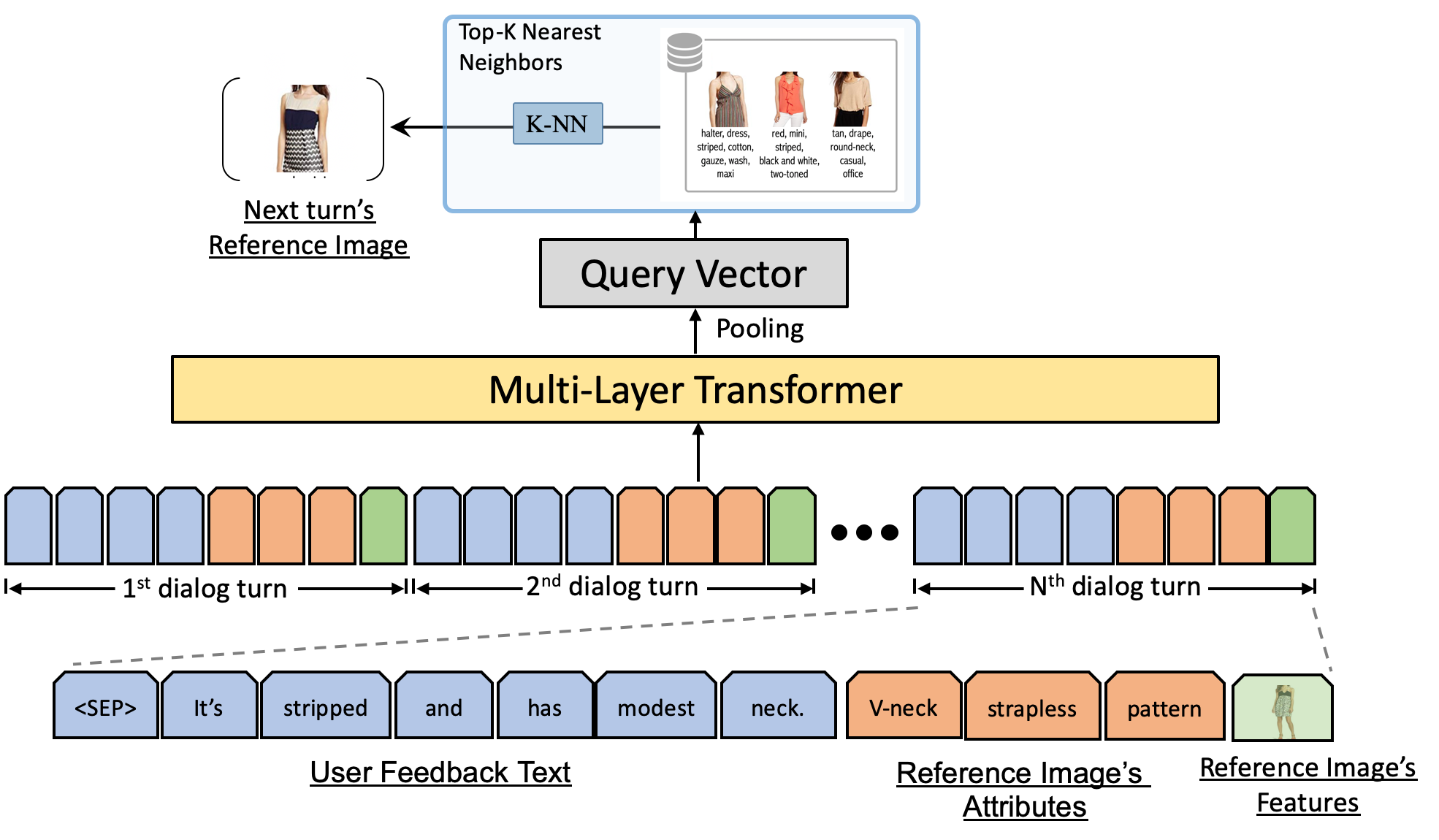}
	\vspace{-1em}
	\caption{Our multimodal transformer model for image retrieval, which integrates, through self-attention, visual attributes with image features, user feedback, and the entire dialog history during each turn, in order to retrieve the next candidate image.\label{fig:tf_ret}}
	\vspace{-1em}
\end{figure}

During training, our dialog-based retrieval model leverages the previously introduced relative captioning model to simulate the user's input at the start of each cycle of the interaction. 
More specifically, the user model is used to generate relative captions for image pairs that occur during each  interaction (which are generally not present in the training data of the captioning model), and enables efficient training of the  interactive retriever without a human in the loop as was done in \cite{guo2018dialog}. For commercial applications, this learning procedure would serve as pre-training to bootstrap and then boost system performance, as it is fine-tuned on real multi-turn interaction data that becomes available. 
The relative captioning model provides the dialog-based retriever at each iteration $j$ with a relative description of the differences between the retrieved image $I_j$ and the target image $I_t$.
Note that only the user model $f_R$ has access to $I_t$, and $f_R$ communicates to the dialog model $f_D$ only through natural language.
Furthermore, to prevent $f_R$ and $f_D$ from developing a coded language among themselves, we pre-train $f_R$ separately on relative captions, and freeze the model parameters when training $f_D$. 

To that end, at each iteration $j$ of the dialog with the user, $f_D$ receives the user model's relative feedback $\{w_i\}^j=f_R(I_j, I_t)$, the top $N$ attributes from $I_j$, and image features of $I_j$ (see Figure~\ref{fig:tf_ret}).
The model attends to these features with a multi-layer transformer to produce a query vector $q_{j}=f_D(\{\{w_i\}^k, f_W(\{a_i\}^k), f_I(I_k)\}_{k=1}^j)$, where $j$ is the current iteration.
The query $q_j$ is used to search the database for the best matching garment based on the Euclidean distance in image feature vector space, and the image of the top result $I_{j+1}$ is returned to the user for the next iteration. \vspace{-0.05in}
\section{Experiments\label{sec:exp}}
We evaluate our multimodal transformer models on the user simulation and interactive fashion retrieval tasks of Fashion IQ. 
We compare against the state-of-the-art hierarchical RNN-based approach from \cite{guo2018dialog} and demonstrate the benefit of the design choices of our baselines and the newly introduced attributes in boosting performance.  
All models are evaluated on the three fashion categories: Dresses, Shirts and Tops\&Tees, following the same data split shown in Table~\ref{table:statistics}. These models establish formal performance benchmarks for the user modeling and dialog-based retrieval tasks of Fashion IQ, and outperform those of (\cite{guo2018dialog}), even when not leveraging attributes as side information (cf. Tables \ref{table:retrieve}, \ref {table:caption}). 

\subsection{Experiment Setup}\label{sec:exp_setup} 

{\bf \noindent Image Features.} 
We realize the image encoder $f_I$ by utilizing an EfficientNet-b7~\cite{tan2019efficientnet} pretrained on the attribute prediction task from the DeepFashion dataset. 
The feature map after the last average pooling layer is flattened to vector of size 2560 which is used as the image representation.  

\begin{table*}[h!]
\begin{center}
\scalebox{0.85}{
\begin{tabular}{lccccccccc}
\toprule
                        & \multicolumn{3}{c}{Dialog Turn 1}  &  \multicolumn{3}{c}{{Dialog Turn 3}} & \multicolumn{3}{c}{{Dialog Turn 5}}   \\
                        & \textbf{P}        & \textbf{R@10} & \textbf{R@50}     & \textbf{P}        & \textbf{R@10} & \textbf{R@50}     & \textbf{P}        & \textbf{R@10} & \textbf{R@50}   \\
\midrule
\multicolumn{7}{l}{\textbf{Dresses}}\\
Guo et al.~\cite{guo2018dialog}               & 89.45              & 6.25 & 20.26                        & 97.49             & 26.95 & 57.78                       & 98.56             & 39.12 & 72.21\\
Ours             & 93.14              & 12.45 & 35.21                       & 97.96             & 36.48 & 68.13                       & 98.39             & 41.35 & 73.63\\
Ours w/ Attr.    & \textbf{93.50}     & \textbf{13.39} & \textbf{35.56}     & \textbf{98.30}    & \textbf{40.11} & \textbf{72.14}     & \textbf{98.69}    & \textbf{46.28} & \textbf{77.24}\\
\midrule
\multicolumn{7}{l}{\textbf{Shirts}}\\
Guo et al.~\cite{guo2018dialog}              & 89.39              & 3.86 & 13.95                        & 97.40             & 21.78  & 47.92                       & 98.48             & 32.94 & 62.03 \\
Ours             & 92.75              & \textbf{11.05} & 28.99              & 98.03             & 30.34 & \textbf{60.32}              & 98.28             & \textbf{33.91} & 63.42 \\
Ours w/Attr.     & \textbf{92.92}     & 11.03 & \textbf{29.03}              & \textbf{98.09}    & \textbf{30.63} & 60.20              & \textbf{98.46}    & 33.69 & \textbf{64.60}\\
\midrule
\multicolumn{7}{l}{\textbf{Tops\&Tees}}\\
Guo et al.~\cite{guo2018dialog}               & 87.89             & 3.03 & 12.34                         & 96.82             & 17.30 & 42.87                       & 98.30             & 29.59 & 60.82\\
Ours             &  93.03            & 11.24 & 30.45                        & 97.88             & 30.22 & 59.95                       & 98.22             & 33.52 & 63.85 \\
Ours w/ Attr.    & \textbf{93.25}    & \textbf{11.74} & \textbf{31.52}      & \textbf{98.10}    & \textbf{31.36} & \textbf{61.76}     & \textbf{98.44}    & \textbf{35.94} & \textbf{66.56}\\
\bottomrule
\end{tabular}
}
\end{center}
\vspace{-1.5em}
\caption{
    \textbf{Dialog-based Image Retrieval.} We report the performance on ranking percentile (P) and recall at N (R@N) at the $1$st, $3$rd and $5$th dialog turns. 
    \label{table:retrieve}
    }
\vspace{-1em}
\end{table*}

\vspace{0.4em}
{\bf \noindent Attribute Prediction.} 
For the attribute model $f_A$, we fine-tune the last linear layer of the previous EfficientNet-b7 using the attribute labels from our Fashion IQ dataset.
Then, we use the fine-tuned EfficientNet-b7 to generate the top-8 attributes for the garment images.

\vspace{0.4em}
{\bf \noindent Attribute and Word Embedding.}
For $f_W$, we use randomly initialized embeddings and they are optimized end-to-end with other components. We use GloVe~\cite{pennington2014}
to encode user feedback words in the retriever, which is pre-trained on an external text corpus to represent each word with a 300-dimensional vector. 

\vspace{0.4em}
{\bf \noindent Transformer Details.} 
The multimodal retrieval model is a 6-layer transformer (256 hidden units, 8 attention heads)\footnote{Our transformer implementation is based on the Harvard NLP library (https://nlp.seas.harvard.edu/2018/04/03/attention.html).}. 
The user's feedback text is padded to a fixed length of 8. 
The transformer's output representations are then pooled and linearly transformed to form the query vector. 
All other parameters are set to their default values. 
The multimodal captioning model has 6 encoding and 6 decoding transformer layers and its caption output is set to maximum word length of 8.
The captioner's loss function is the cross entropy and the retrieval's is the triplet-based loss as defined in ~\cite{guo2018dialog}. For further details regarding model training please consult Appendix~\ref{app:retrieve_results}. We will also make our source code available.\footnote{https://github.com/XiaoxiaoGuo/fashion-iq}  

\begin{table}[t!]
\begin{center}
\scalebox{0.85}{
\begin{tabular}{lcccc}
\toprule
                     & \textbf{BLEU-4}   & \textbf{Rouge-L}  & \textbf{CIDEr}    & \textbf{SPICE} \\
\midrule
\multicolumn{5}{l}{\textbf{Dresses}} \\
Guo et al.~\cite{guo2018dialog}      & 17.4              & 53.6              & 48.9              & 32.1\\
Ours                         & 20.7              & 56.3              & 78.5              & 34.4 \\
Ours w/ Attr.                & \textbf{21.1}     & \textbf{57.1}     & \textbf{80.6}     & \textbf{36.1} \\
\midrule
\multicolumn{5}{l}{\textbf{Shirts}}\\
Guo et al.~\cite{guo2018dialog}      & 19.6              & 53.8              & 52.6              & 32.0\\
Ours                         & 22.3              & 56.4              & 84.1              & 34.7 \\
Ours w/ Attr.                & \textbf{24.2}     & \textbf{57.5}     & \textbf{92.1}     & \textbf{35.4} \\
\midrule
\multicolumn{5}{l}{\textbf{Tops\&Tees}}\\
Guo et al.~\cite{guo2018dialog}      & 15.7              & 50.5              & 41.1              & 30.6 \\
Ours                         & 20.6              & 54.8              & 79.8              & \textbf{36.4} \\
Ours w/ Attr.                & \textbf{22.1}     & \textbf{55.4}     & \textbf{82.3}     & 35.0 \\
\bottomrule
\end{tabular}
}
\end{center}
\vspace{-1.5em}
\caption{
    \textbf{Relative Captioning.} 
    Our multimodal transformer captioning model outperforms the state-of-the-art RNN-based approach \cite{guo2018dialog} on standard image captioning metrics across all datasets. 
\label{table:caption}}
\vspace{-1em}
\end{table}
\subsection{Experimental  Results}\label{sec:exp_results}

{\bf \noindent Relative Captioning.} 
Table~\ref{table:caption} summarizes the performance of our multimodal transformer approach compared to the RNN-based approach from~\cite{guo2018dialog}. 
Our transformer method outperforms the RNN-based baseline across all metrics. 
Moreover, the attribute-aware transformer model improves over the attribute-agnostic variant, suggesting that attribute information is complementary to the raw visual signals and improves relative captioning performance. 

{\bf \noindent Dialog-based Image Retrieval.}
To test dialog-based retrieval performance, we paired each retrieval model with user models and ran the dialog interaction for five turns, starting from a random test image, to retrieve each target test image.
Note that the user simulator and the retriever are trained independently, and can communicate only via generated captions and retrieved images.
Image retrieval performance is quantified by the average ranking percentile of the target image on the test data split and the recall of the target image at top-N (R@N) in Table~\ref{table:retrieve}. 
Our transformer-based models outperform the previous RNN-based SOTA by a significant margin.
In addition, the attribute-aware model  \KG{produces} better retrieval results overall,  suggesting that the newly introduced attributes in our dataset are of benefit to the ``downstream" dialog-based retrieval task. Additional ablations and visualization examples are in Appendix~\ref{app:retrieve_results}. 

\section{Conclusions}
\vspace{-2mm}
We introduced Fashion IQ, a new dataset for research on natural language based
image retrieval systems, which is situated in the detail-critical fashion domain.
Fashion IQ is the first product-oriented dataset that makes available both high-quality, human-annotated relative captions, and image attributes derived from product descriptions. We showed that image attributes and natural language feedback are complementary to each other, and that combining them \KG{leads} to significant improvements
\KG{to} 
interactive image retrieval systems.
The natural language interface investigated in this paper overcomes the need to engineer brittle and cumbersome ontologies for every new application, and provides a more natural and expressive way for users to compose novel and complex queries, compared to structured interfaces.
We believe that both the dataset and the frameworks explored in this paper will serve as important stepping stones toward building ever more effective interactive image retrieval systems in the future.

{\small
\bibliographystyle{ieee_fullname.bst}
\bibliography{egbib.bib}

\begin{thebibliography}{10}\itemsep=-1pt

\bibitem{al2017fashion}
Ziad Al-Halah, Rainer Stiefelhagen, and Kristen Grauman.
\newblock Fashion forward: Forecasting visual style in fashion.
\newblock In {\em ICCV}, 2017.

\bibitem{VQA}
Stanislaw Antol, Aishwarya Agrawal, Jiasen Lu, Margaret Mitchell, Dhruv Batra,
  C.~Lawrence Zitnick, and Devi Parikh.
\newblock {VQA}: {V}isual {Q}uestion {A}nswering.
\newblock In {\em ICCV}, 2015.

\bibitem{barbu2013saying}
Daniel Barrett, Andrei Barbu, N Siddharth, and Jeffrey~Mark Siskind.
\newblock Saying what you're looking for: Linguistics meets video search.
\newblock {\em IEEE Transactions on Pattern Analysis and Machine Intelligence},
  38(10), 2016.

\bibitem{berg2010automatic}
Tamara~L Berg, Alexander~C Berg, and Jonathan Shih.
\newblock Automatic attribute discovery and characterization from noisy web
  data.
\newblock In {\em ECCV}, 2010.

\bibitem{budzianowski2018}
Pawe{\l} Budzianowski, Tsung-Hsien Wen, Bo-Hsiang Tseng, I{\~n}igo Casanueva,
  Stefan Ultes, Osman Ramadan, and Milica Ga{\v{s}}i{\'c}.
\newblock {M}ulti{WOZ} - a large-scale multi-domain {W}izard-of-{O}z dataset
  for task-oriented dialogue modelling.
\newblock In {\em EMNLP}, 2018.

\bibitem{chen2015deep}
Qiang Chen, Junshi Huang, Rogerio Feris, Lisa~M Brown, Jian Dong, and Shuicheng
  Yan.
\newblock Deep domain adaptation for describing people based on fine-grained
  clothing attributes.
\newblock In {\em CVPR}, 2015.

\bibitem{cheneccv2020}
Y. Chen and L. Bazzani.
\newblock Learning joint visual semantic matching embeddings for
  language-guided retrieval.
\newblock In {\em ECCV}, 2020.

\bibitem{chen2020image}
Yanbei Chen, Shaogang Gong, and Loris Bazzani.
\newblock Image search with text feedback by visiolinguistic attention
  learning.
\newblock In {\em Proceedings of the IEEE/CVF Conference on Computer Vision and
  Pattern Recognition}, pages 3001--3011, 2020.

\bibitem{das2018embodied}
Abhishek Das, Samyak Datta, Georgia Gkioxari, Stefan Lee, Devi Parikh, and
  Dhruv Batra.
\newblock Embodied question answering.
\newblock In {\em CVPR}, 2018.

\bibitem{das2016visual}
Abhishek Das, Satwik Kottur, Khushi Gupta, Avi Singh, Deshraj Yadav,
  Jos{\'e}~MF Moura, Devi Parikh, and Dhruv Batra.
\newblock Visual dialog.
\newblock In {\em CVPR}, 2017.

\bibitem{das2017learning}
Abhishek Das, Satwik Kottur, Jos{\'e}~MF Moura, Stefan Lee, and Dhruv Batra.
\newblock Learning cooperative visual dialog agents with deep reinforcement
  learning.
\newblock In {\em ICCV}, 2017.

\bibitem{de2016guesswhat}
Harm de Vries, Florian Strub, Sarath Chandar, Olivier Pietquin, Hugo
  Larochelle, and Aaron Courville.
\newblock Guesswhat?! visual object discovery through multi-modal dialogue.
\newblock In {\em CVPR}, 2017.

\bibitem{dodds2020modality}
Eric Dodds, Jack Culpepper, Simao Herdade, Yang Zhang, and Kofi Boakye.
\newblock Modality-agnostic attention fusion for visual search with text
  feedback.
\newblock {\em arXiv preprint arXiv:2007.00145}, 2020.

\bibitem{forbes2019neural}
Maxwell Forbes, Christine Kaeser-Chen, Piyush Sharma, and Serge Belongie.
\newblock Neural naturalist: Generating fine-grained image comparisons.
\newblock In {\em Conference on Empirical Methods in Natural Language
  Processing (EMNLP)}, Hong Kong, 2019.

\bibitem{ge2019deepfashion2}
Yuying Ge, Ruimao Zhang, Lingyun Wu, Xiaogang Wang, Xiaoou Tang, and Ping Luo.
\newblock Deepfashion2: A versatile benchmark for detection, pose estimation,
  segmentation and re-identification of clothing images.
\newblock {\em arXiv preprint arXiv:1901.07973}, 2019.

\bibitem{guo2019imaterialist}
Sheng Guo, Weilin Huang, Xiao Zhang, Prasanna Srikhanta, Yin Cui, Yuan Li,
  Hartwig Adam, Matthew~R Scott, and Serge Belongie.
\newblock The imaterialist fashion attribute dataset.
\newblock In {\em CVPR Workshop on Computer Vision for Fashion, Art and
  Design}, 2019.

\bibitem{guo2018dialog}
Xiaoxiao Guo, Hui Wu, Yu Cheng, Steven Rennie, Gerald Tesauro, and Rogerio
  Feris.
\newblock Dialog-based interactive image retrieval.
\newblock In {\em NeurIPS}, 2018.

\bibitem{hadi2015buy}
M Hadi~Kiapour, Xufeng Han, Svetlana Lazebnik, Alexander~C Berg, and Tamara~L
  Berg.
\newblock Where to buy it: Matching street clothing photos in online shops.
\newblock In {\em ICCV}, 2015.

\bibitem{han2017automatic}
Xintong Han, Zuxuan Wu, Phoenix~X Huang, Xiao Zhang, Menglong Zhu, Yuan Li,
  Yang Zhao, and Larry~S Davis.
\newblock Automatic spatially-aware fashion concept discovery.
\newblock In {\em ICCV}, 2017.

\bibitem{he2016ups}
Ruining He and Julian McAuley.
\newblock Ups and downs: Modeling the visual evolution of fashion trends with
  one-class collaborative filtering.
\newblock In {\em WWW}, 2016.

\bibitem{hsiao2017learning}
Wei-Lin Hsiao and Kristen Grauman.
\newblock Learning the latent “look”: Unsupervised discovery of a
  style-coherent embedding from fashion images.
\newblock In {\em ICCV}, 2017.

\bibitem{hsiao2018creating}
Wei-Lin Hsiao and Kristen Grauman.
\newblock Creating capsule wardrobes from fashion images.
\newblock In {\em PCVPR}, 2018.

\bibitem{hsiao2020vibe}
Wei-Lin Hsiao and Kristen Grauman.
\newblock Vibe: Dressing for diverse body shapes.
\newblock In {\em CVPR}, 2020.

\bibitem{huang2015cross}
Junshi Huang, Rogerio~S Feris, Qiang Chen, and Shuicheng Yan.
\newblock Cross-domain image retrieval with a dual attribute-aware ranking
  network.
\newblock In {\em ICCV}, 2015.

\bibitem{jhamtani2018learning}
Harsh Jhamtani and Taylor Berg-Kirkpatrick.
\newblock Learning to describe differences between pairs of similar images.
\newblock In {\em Proceedings of the 2018 Conference on Empirical Methods in
  Natural Language Processing}, pages 4024--4034, 2018.

\bibitem{jia2020fashionpedia}
Menglin Jia, Mengyun Shi, Mikhail Sirotenko, Yin Cui, Claire Cardie, Bharath
  Hariharan, Hartwig Adam, and Serge Belongie.
\newblock Fashionpedia: Ontology, segmentation, and an attribute localization
  dataset.
\newblock In {\em ECCV}, 2020.

\bibitem{kiapour2014hipster}
M~Hadi Kiapour, Kota Yamaguchi, Alexander~C Berg, and Tamara~L Berg.
\newblock Hipster wars: Discovering elements of fashion styles.
\newblock In {\em ECCV}, 2014.

\bibitem{kiros2014unifying}
Ryan Kiros, Ruslan Salakhutdinov, and Richard~S Zemel.
\newblock Unifying visual-semantic embeddings with multimodal neural language
  models.
\newblock {\em arXiv preprint arXiv:1411.2539}, 2014.

\bibitem{kovashka2013attribute}
Adriana Kovashka and Kristen Grauman.
\newblock Attribute pivots for guiding relevance feedback in image search.
\newblock In {\em ICCV}, 2013.

\bibitem{shades2014}
A. Kovashka and K. Grauman.
\newblock Discovering shades of attribute meaning with the crowd.
\newblock In {\em ECCV Workshop on Parts and Attributes}, 2014.

\bibitem{kovashka2017attributes}
Adriana Kovashka and Kristen Grauman.
\newblock Attributes for image retrieval.
\newblock In {\em Visual Attributes}. Springer, 2017.

\bibitem{kovashka2012}
Adriana Kovashka, Devi Parikh, and Kristen Grauman.
\newblock Whittlesearch: Image search with relative attribute feedback.
\newblock In {\em CVPR}, 2012.

\bibitem{laenen2017cross}
Katrien Laenen, Susana Zoghbi, and Marie-Francine Moens.
\newblock Cross-modal search for fashion attributes.
\newblock In {\em KDD Workshop on Machine Learning Meets Fashion}, 2017.

\bibitem{li2018towards}
Raymond Li, Samira~Ebrahimi Kahou, Hannes Schulz, Vincent Michalski, Laurent
  Charlin, and Chris Pal.
\newblock Towards deep conversational recommendations.
\newblock In {\em NeurIPS}, 2018.

\bibitem{li2017person}
Shuang Li, Tong Xiao, Hongsheng Li, Bolei Zhou, Dayu Yue, and Xiaogang Wang.
\newblock Person search with natural language description.
\newblock In {\em CVPR}, 2017.

\bibitem{li2020oscar}
Xiujun Li, Xi Yin, Chunyuan Li, Pengchuan Zhang, Xiaowei Hu, Lei Zhang, Lijuan
  Wang, Houdong Hu, Li Dong, Furu Wei, et~al.
\newblock Oscar: Object-semantics aligned pre-training for vision-language
  tasks.
\newblock In {\em European Conference on Computer Vision}, 2020.

\bibitem{DeepFashion}
Ziwei Liu, Ping Luo, Shi Qiu, Xiaogang Wang, and Xiaoou Tang.
\newblock Deepfashion: Powering robust clothes recognition and retrieval with
  rich annotations.
\newblock In {\em CVPR}, 2016.

\bibitem{lu2019vilbert}
Jiasen Lu, Dhruv Batra, Devi Parikh, and Stefan Lee.
\newblock Vilbert: Pretraining task-agnostic visiolinguistic representations
  for vision-and-language tasks.
\newblock In {\em NeurIPS}, 2019.

\bibitem{lu2017fully}
Yongxi Lu, Abhishek Kumar, Shuangfei Zhai, Yu Cheng, Tara Javidi, and Rogerio
  Feris.
\newblock Fully-adaptive feature sharing in multi-task networks with
  applications in person attribute classification.
\newblock In {\em CVPR}, 2017.

\bibitem{mcauley2015image}
Julian McAuley, Christopher Targett, Qinfeng Shi, and Anton Van Den~Hengel.
\newblock Image-based recommendations on styles and substitutes.
\newblock In {\em SIGIR}, 2015.

\bibitem{parikh2011relative}
Devi Parikh and Kristen Grauman.
\newblock Relative attributes.
\newblock In {\em ICCV}, 2011.

\bibitem{park2019robust}
Dong~Huk Park, Trevor Darrell, and Anna Rohrbach.
\newblock Robust change captioning.
\newblock In {\em Proceedings of the IEEE International Conference on Computer
  Vision}, pages 4624--4633, 2019.

\bibitem{pennington2014}
Jeffrey Pennington, Richard Socher, and Christopher~D. Manning.
\newblock Glove: Global vectors for word representation.
\newblock In {\em Empirical Methods in Natural Language Processing (EMNLP)},
  2014.

\bibitem{plummer2019give}
Bryan Plummer, Hadi Kiapour, Shuai Zheng, and Robinson Piramuthu.
\newblock Give me a hint! navigating image databases using human-in-the-loop
  feedback.
\newblock In {\em WACV}, 2019.

\bibitem{rennie2016self}
Steven~J Rennie, Etienne Marcheret, Youssef Mroueh, Jarret Ross, and Vaibhava
  Goel.
\newblock Self-critical sequence training for image captioning.
\newblock In {\em CVPR}, 2017.

\bibitem{rostamzadeh2018fashion}
Negar Rostamzadeh, Seyedarian Hosseini, Thomas Boquet, Wojciech Stokowiec, Ying
  Zhang, Christian Jauvin, and Chris Pal.
\newblock Fashion-gen: The generative fashion dataset and challenge.
\newblock {\em arXiv preprint arXiv:1806.08317}, 2018.

\bibitem{rui1998relevance}
Yong Rui, Thomas~S Huang, Michael Ortega, and Sharad Mehrotra.
\newblock Relevance feedback: a power tool for interactive content-based image
  retrieval.
\newblock {\em IEEE Transactions on circuits and systems for video technology},
  8(5):644--655, 1998.

\bibitem{saha2018towards}
Amrita Saha, Mitesh~M Khapra, and Karthik Sankaranarayanan.
\newblock Towards building large scale multimodal domain-aware conversation
  systems.
\newblock In {\em AAAI}, 2018.

\bibitem{sharmanska2013learning}
Viktoriia Sharmanska, Novi Quadrianto, and Christoph~H Lampert.
\newblock Learning to rank using privileged information.
\newblock In {\em ICCV}, 2013.

\bibitem{simo2015neuroaesthetics}
Edgar Simo-Serra, Sanja Fidler, Francesc Moreno-Noguer, and Raquel Urtasun.
\newblock Neuroaesthetics in fashion: Modeling the perception of
  fashionability.
\newblock In {\em CVPR}, 2015.

\bibitem{simo2016fashion}
Edgar Simo-Serra and Hiroshi Ishikawa.
\newblock Fashion style in 128 floats: Joint ranking and classification using
  weak data for feature extraction.
\newblock In {\em CVPR}, 2016.

\bibitem{souri2016deep}
Yaser Souri, Erfan Noury, and Ehsan Adeli.
\newblock Deep relative attributes.
\newblock In {\em ACCV}, 2016.

\bibitem{strub2017end}
Florian Strub, Harm de Vries, Jeremie Mary, Bilal Piot, Aaron Courville, and
  Olivier Pietquin.
\newblock End-to-end optimization of goal-driven and visually grounded dialogue
  systems.
\newblock In {\em IJCAI}, 2017.

\bibitem{sun2019videobert}
Chen Sun, Austin Myers, Carl Vondrick, Kevin Murphy, and Cordelia Schmid.
\newblock Videobert: A joint model for video and language representation
  learning.
\newblock In {\em ICCV}, 2019.

\bibitem{tan2019}
Fuwen Tan, Paola Cascante-Bonilla, Xiaoxiao Guo, Steven Wu, Gerald Hui, Song
  Feng, and Vicente Ordonez.
\newblock Drill-down: Interactive retrieval of complex scenes using natural
  language queries.
\newblock In {\em NeurIPS}, 2019.

\bibitem{tan2019lxmert}
Hao Tan and Mohit Bansal.
\newblock Lxmert: Learning cross-modality encoder representations from
  transformers.
\newblock In {\em EMNLP}, 2019.

\bibitem{tan2019expressing}
Hao Tan, Franck Dernoncourt, Zhe Lin, Trung Bui, and Mohit Bansal.
\newblock Expressing visual relationships via language.
\newblock {\em arXiv preprint arXiv:1906.07689}, 2019.

\bibitem{tan2019efficientnet}
Mingxing Tan and Quoc Le.
\newblock Efficientnet: Rethinking model scaling for convolutional neural
  networks.
\newblock In {\em International Conference on Machine Learning}, pages
  6105--6114, 2019.

\bibitem{tapaswi2016movieqa}
Makarand Tapaswi, Yukun Zhu, Rainer Stiefelhagen, Antonio Torralba, Raquel
  Urtasun, and Sanja Fidler.
\newblock Movieqa: Understanding stories in movies through question-answering.
\newblock In {\em CVPR}, 2016.

\bibitem{tellex2009towards}
Stefanie Tellex and Deb Roy.
\newblock Towards surveillance video search by natural language query.
\newblock In {\em ACM International Conference on Image and Video Retrieval},
  2009.

\bibitem{vapnik2009new}
Vladimir Vapnik and Akshay Vashist.
\newblock A new learning paradigm: Learning using privileged information.
\newblock {\em Neural networks}, 22(5-6):544--557, 2009.

\bibitem{vaswani2017attention}
Ashish Vaswani, Noam Shazeer, Niki Parmar, Jakob Uszkoreit, Llion Jones,
  Aidan~N Gomez, {\L}ukasz Kaiser, and Illia Polosukhin.
\newblock Attention is all you need.
\newblock In {\em Advances in neural information processing systems}, 2017.

\bibitem{veit2015learning}
Andreas Veit, Balazs Kovacs, Sean Bell, Julian McAuley, Kavita Bala, and Serge
  Belongie.
\newblock Learning visual clothing style with heterogeneous dyadic
  co-occurrences.
\newblock In {\em ICCV}, 2015.

\bibitem{vinyals2015show}
Oriol Vinyals, Alexander Toshev, Samy Bengio, and Dumitru Erhan.
\newblock Show and tell: A neural image caption generator.
\newblock In {\em CVPR}, 2015.

\bibitem{vo2018composing}
Nam Vo, Lu Jiang, Chen Sun, Kevin Murphy, Li-Jia Li, Li Fei-Fei, and James
  Hays.
\newblock Composing text and image for image retrieval-an empirical odyssey.
\newblock In {\em CVPR}, 2019.

\bibitem{wu2018image}
Qi Wu, Chunhua Shen, Peng Wang, Anthony Dick, and Anton van~den Hengel.
\newblock Image captioning and visual question answering based on attributes
  and external knowledge.
\newblock {\em IEEE transactions on pattern analysis and machine intelligence},
  40(6):1367--1381, 2018.

\bibitem{icml2015_xuc15}
Kelvin Xu, Jimmy Ba, Ryan Kiros, Kyunghyun Cho, Aaron Courville, Ruslan
  Salakhudinov, Rich Zemel, and Yoshua Bengio.
\newblock Show, attend and tell: Neural image caption generation with visual
  attention.
\newblock In {\em ICML}, 2015.

\bibitem{yang2017visual}
Fan Yang, Ajinkya Kale, Yury Bubnov, Leon Stein, Qiaosong Wang, Hadi Kiapour,
  and Robinson Piramuthu.
\newblock Visual search at ebay.
\newblock In {\em KDD}, 2017.

\bibitem{yang2014clothing}
Wei Yang, Ping Luo, and Liang Lin.
\newblock Clothing co-parsing by joint image segmentation and labeling.
\newblock In {\em CVPR}, 2014.

\bibitem{yao2017boosting}
Ting Yao, Yingwei Pan, Yehao Li, Zhaofan Qiu, and Tao Mei.
\newblock Boosting image captioning with attributes.
\newblock In {\em Proceedings of the IEEE International Conference on Computer
  Vision}, 2017.

\bibitem{you2016image}
Quanzeng You, Hailin Jin, Zhaowen Wang, Chen Fang, and Jiebo Luo.
\newblock Image captioning with semantic attention.
\newblock In {\em Proceedings of the IEEE conference on computer vision and
  pattern recognition}, pages 4651--4659, 2016.

\bibitem{yu2014fine}
Aron Yu and Kristen Grauman.
\newblock Fine-grained visual comparisons with local learning.
\newblock In {\em CVPR}, 2014.

\bibitem{zhao2017memory}
Bo Zhao, Jiashi Feng, Xiao Wu, and Shuicheng Yan.
\newblock Memory-augmented attribute manipulation networks for interactive
  fashion search.
\newblock In {\em CVPR}, 2017.

\bibitem{zheng2018modanet}
Shuai Zheng, Fan Yang, M~Hadi Kiapour, and Robinson Piramuthu.
\newblock Modanet: A large-scale street fashion dataset with polygon
  annotations.
\newblock {\em arXiv preprint arXiv:1807.01394}, 2018.

\bibitem{zhou2020unified}
Luowei Zhou, Hamid Palangi, Lei Zhang, Houdong Hu, Jason~J Corso, and Jianfeng
  Gao.
\newblock Unified vision-language pre-training for image captioning and vqa.
\newblock In {\em AAAI}, 2020.

\end{thebibliography}
}

\newpage
\clearpage
\begin{center}
{\bf {\Large Appendix\\} }
\end{center}
\appendix
\setcounter{page}{1}

\section{Additional Information on Fashion IQ \label{app:attribute_prediction}}

\begin{figure}[h!]
\begin{center}
\includegraphics[width=1\linewidth]{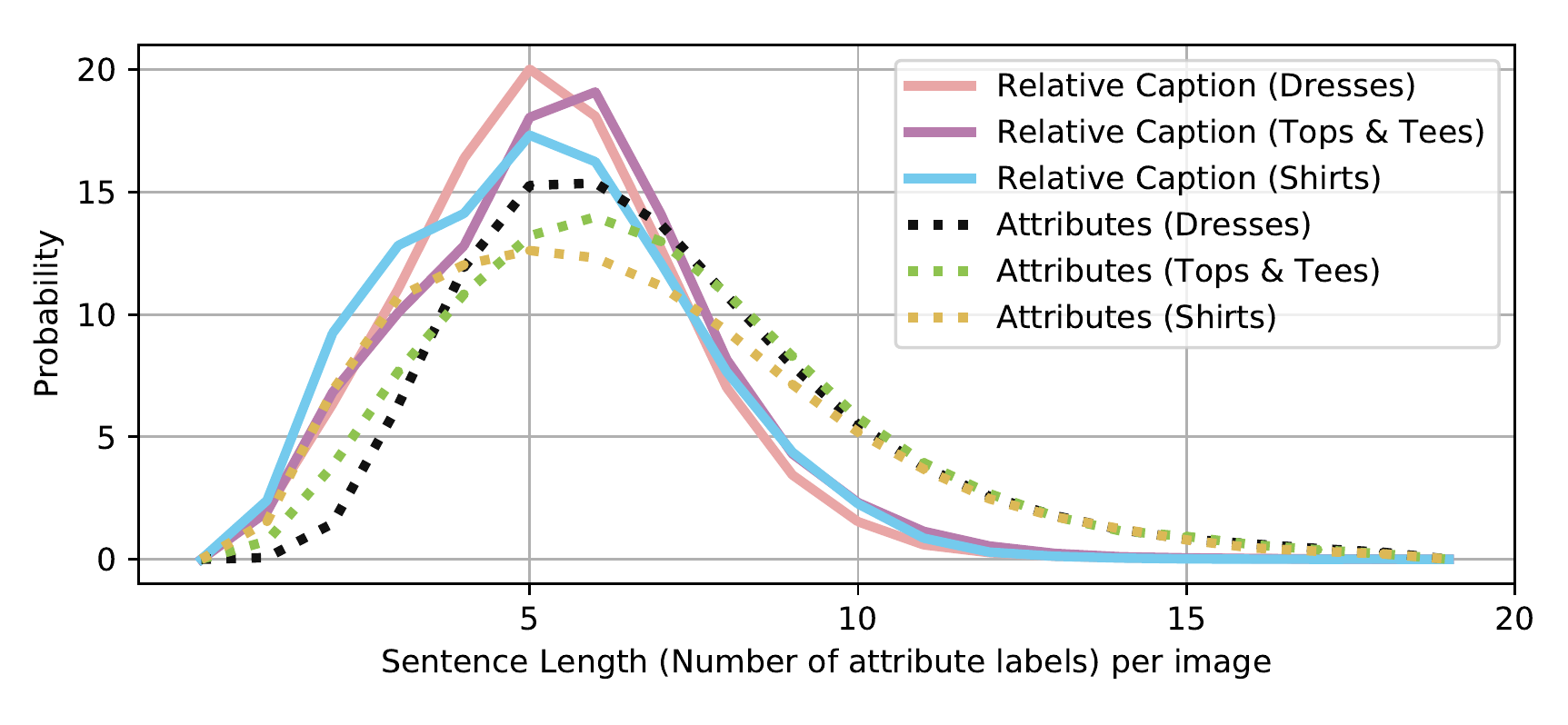}
\end{center}
\vspace{-1.5em}
\caption{Distribution of sentence lengths and number of attribute labels per image.
\label{fig:distribution}
  }
\vspace{-1em}
\end{figure}

\begin{figure*}[t!]
	\centering
	\includegraphics[width=.8\linewidth]{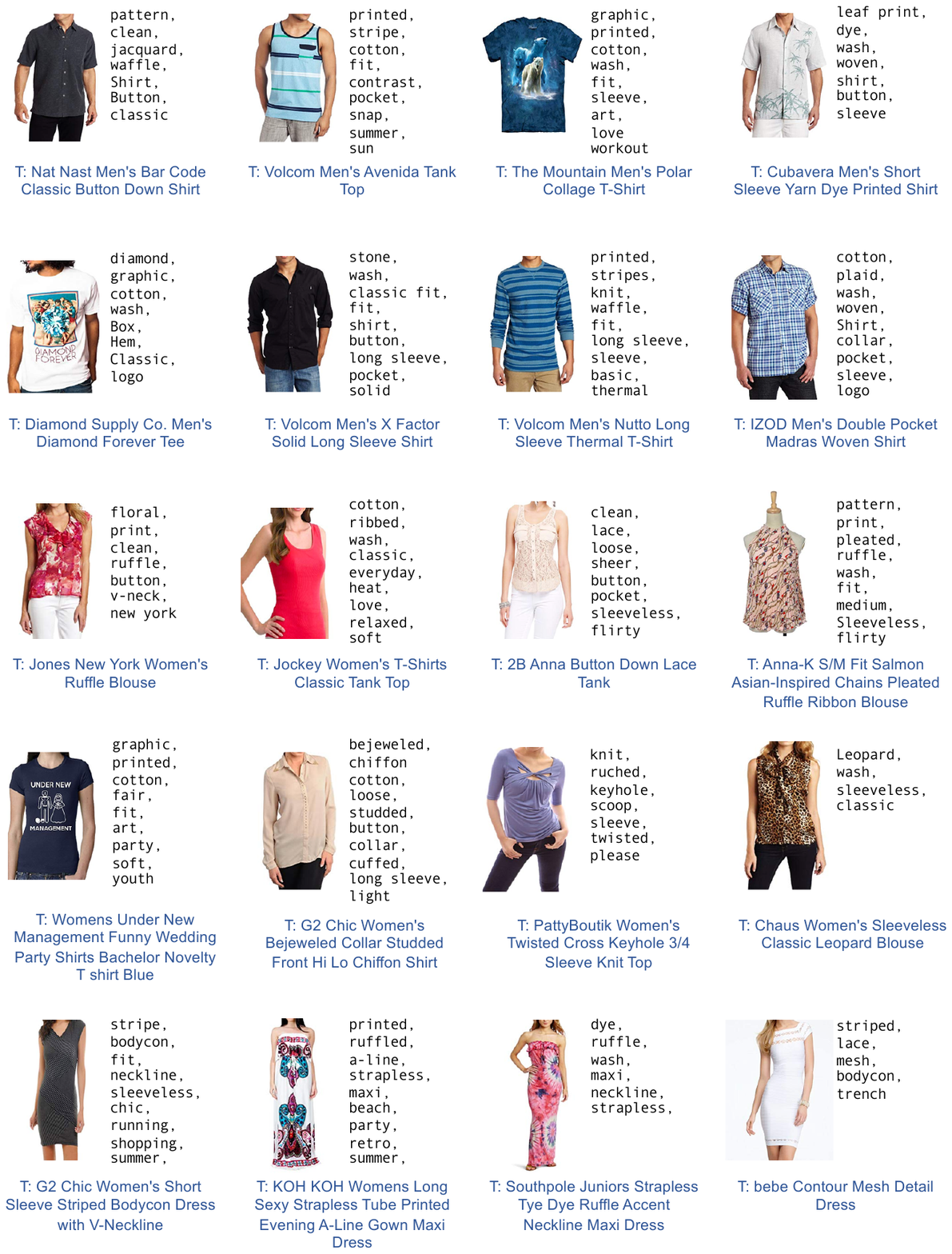}
	\caption{Examples of the original product title descriptions (T) and the collected
	attribute labels (on the right of each image). }
	\label{fig:amazon2}
\end{figure*}

Our dataset is publicly available and free for academic use.\footnote{https://github.com/XiaoxiaoGuo/fashion-iq} Figure~\ref{fig:distribution} depicts the empirical distributions of relative caption length and number of
attributes per image for all subsets of Fashion IQ.
In Figure~\ref{fig:amazon2}, we show more examples of the original product
titles and the derived attributes. 

\paragraph{Attribute Prediction.} 
The raw attribute labels extracted from the product websites may be noisy or incomplete, therefore, to address this, we utilize the DeepFashion attributes to complete and de-noise the attribute labels in Fashion IQ. 
Specifically, we first train an Attribute Prediction Network, based on the  EfficientNet-b7 architecture,\footnote{https://github.com/lukemelas/EfficientNet-PyTorch} to predict the DeepFashion attributes, using the multi-label binary cross-entropy loss. 
After training on DeepFashion labels, we fine-tune the last layer on each of our Fashion IQ categories (namely Dresses, Shirts, and Tops \& Tees) with the same loss function. The fine-tuning step adjusts the attribute prediction to our categories' attribute distribution. We then use the attribute network to predict the top attribute labels based on their output values. All images have the same number of attribute labels (that is, 8 attributes per image).

\section{Single-turn Image Retrieval}
\label{sec:retrieve}

\begin{table*}[ht]
\begin{center}
\scalebox{0.80}{
\begin{tabular}{c l  c c c }
\toprule
&  & \multicolumn{3}{c}{\textbf{R@10} (\textbf{R@50})} \\
\cline{3-5}
&  & Dresses & Shirts & Tops\&Tees \\
\midrule
\multicolumn{2}{l}{\textbf{Multi-modality retrieval} } &&& \\
A & Image+attributes+relative captions, gating on relative captions.
& 11.24 (\textbf{32.39})  & \textbf{13.73} (\textbf{37.03})  & \textbf{13.52} (\textbf{34.73})   \\
B & Image+relative captions, gating on relative captions.
 & \textbf{11.49} (29.99)  & 13.68 (35.61)  & 11.36 (30.67)  \\
C & Image+relative captions \cite{guo2018dialog}.
 & 10.52 (28.98)  & 13.44 (34.60)  & 11.36 (30.42)  \\\midrule
\multicolumn{2}{l}{\textbf{Single-modality baselines}} &&& \\
D & Relative feedback only.
& 6.94 (23.00)  & 9.24 (27.54)  & 10.02 (26.46)   \\
E & Image feature only.
& 4.20 (13.29)  & 4.51 (14.47)  & 4.13 (14.30)   \\
F & Attribute feature only.
& 2.57 (11.02)  & 4.66 (14.96)  & 4.77 (13.76)     \\
 \bottomrule
\end{tabular}
}
\end{center}
\caption{Results on single-turn image retrieval. \label{table:encoder}}
\end{table*}

As discussed in Sec.~\ref{sec:applications}, we identified three main applications for our Fashion IQ dataset and we demonstrated how the dataset can be leveraged to achieve state-of-the-art performance in relative captioning and dialog-based image retrieval. 
We show here how Fashion IQ can be used in the third task, i.e., single-turn image retrieval.

In this task, given a reference image and a feedback sentence, we aim to retrieve the target image by composing
the two modalities. 
The retrieval experiments use the portion of the dataset that has relative caption annotations. The two relative caption annotations associated with each image are treated as two separate queries during training and testing.
This setting can be thought of as the single-turn scenario in an interactive image retrieval system and has a similar setup as previous work on modifying image query using textual descriptions~\cite{vo2018composing,cheneccv2020,dodds2020modality,chen2020image}.
In this section, we provide empirical studies comparing different combinations of query modalities for
retrieval, including relative feedback, image features, and attribute features.
Specifically, the images were encoded using a pre-trained ResNet-101 network;
the attributes were encoded based on the output of our Attribute Prediction Network; and
the relative feedback sentences were encoded using Gated Recurrent Networks with one hidden layer.
We used pairwise ranking loss~\cite{kiros2014unifying} for all methods with the best margin
parameters for each method selected using the retrieval score on the validation
set. 
We include a baseline model from \cite{guo2018dialog}, which uses the concatenation
of the image feature (after linear embedding) with the encoded relative caption features. We also included two models based on \cite{vo2018composing}, with an additional gating connection, which allows the direct pass of one modality to
the embedding space and has been shown to be effective for jointly
modeling image and text modalities for retrieval. 

We reported the retrieval results on the test set in Table~\ref{table:encoder}.  We found that the best performance was achieved by using all three modalities and applying a gating connection on the encoded natural language feedback (Model A). 
The gating connection on the text feature is shown to be effective for retrieval (comparing B and C), which confirms the informative nature of relative feedback for image retrieval. 
Similar observations can be made in the cases of single-modality studies, where the relative feedback modality (model D) significantly outperformed other modalities (models E and F). 
Finally, Removing attribute features resulted in generally inferior performance (comparing A and B) across the three categories, 
demonstrating the benefit of incorporating attribute labels, concurring with our observations in user modeling experiments and dialog-based retrieval experiments.

\section{Additional Results on Interactive Image Retrieval\label{app:retrieve_results}}

\paragraph{Additional Experimental Details.} To reduce the evaluation variance, we randomly generate a list of initial image pairs (i.e., a target and a reference image), and we evaluated all methods with the same list of the initial image pairs. We use Beam Search with beam size 5 to generate the relative captions as feedback to the retriever model. When training the retriever models, we use greedy decoding for faster training speed. 
The average ranking percentile is computed as $P=\frac{1}{N}\sum_{i=1}^{N}(1-\frac{r_{i}}{N})$, where $r_{i}$ is the ranking  of the $i$-th target and $N$ is the total number of candidates.

\paragraph{Ablative Studies on the Transformer models. } We proposed two Transformer-based models for the interactive image retrieval task, namely the Transformer-based user simulator and the Transformer-based retrieval model. In the ablative studies, we pair the Transformer-based models with the RNN-based counterpart~\cite{guo2018dialog} to assign the improvement credit. Table~\ref{table:additional_retrieval} summarizes the retrieval performance for different combinations. For the same retriever model, the improved user model always improves the retrieval performance for the first turn. As the interaction continues, other factors, including the retrieved image distribution and the simulated feedback diversity, jointly affect the retrieval performance. The improved user model achieved competitive or better scores on average. For the same user model, the Transformer-based retriever model achieved overall better retrieval performance averaged over dialog turns, showing that Transformer-based models effectively aggregate the multimodal information for image retrieval.

\begin{figure}
	\begin{center}
		\includegraphics[width=\linewidth]{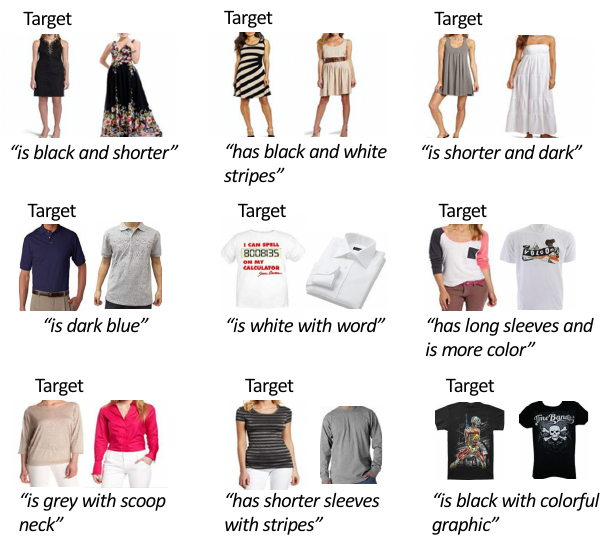}
	\end{center}
	\vspace{-1em}
	\caption{Examples of generated captions from the user model.\label{fig:gencaptions}}
\end{figure}

\begin{figure}
	\begin{center}
		\includegraphics[width=\linewidth]{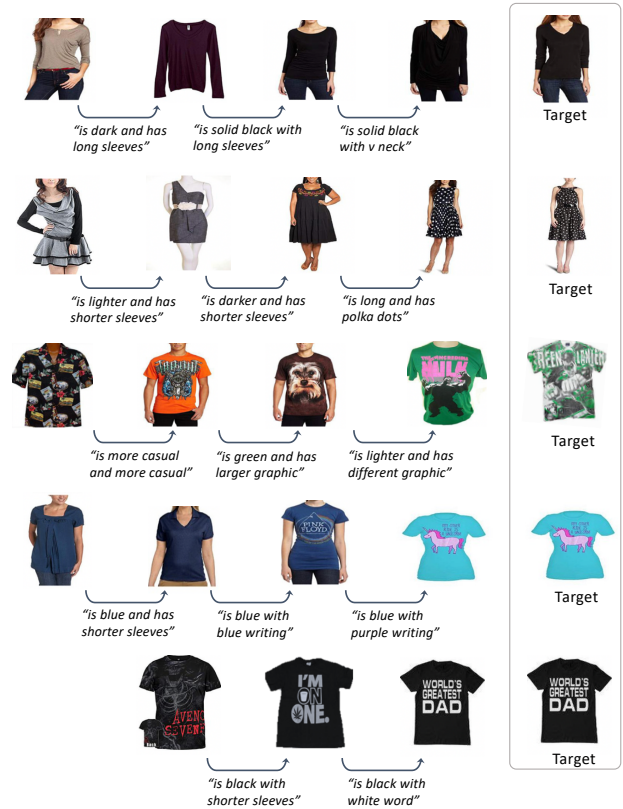}
	\end{center}
	\caption{Examples of the simulator interacting with the dialog manager system. The right-most column shows the target images.\label{fig:more_results2}}
\end{figure}

\paragraph{Visualization.} 
Figure~\ref{fig:gencaptions} shows examples of generated relative captions
from the user model, which contain flexible expressions and both single and composite phrases to describe the differences of the images. Figure~\ref{fig:more_results2} shows examples of the user model interacting with the dialog based retriever. In all examples, the target images reached final rankings within the top 50 images. The target images ranked incrementally higher during the dialog and the candidate images were more visually similar to the target images.
These examples show that the dialog manager is able to refine the candidate selection given the user feedback,
exhibiting promising behavior across different clothing categories.

\begin{table*}[t!]
\begin{center}
\scalebox{0.85}{
\begin{tabular}{lcccccccccccc}
\toprule
& \multicolumn{3}{c}{Dialog Turn 1}  &  \multicolumn{3}{c}{{Dialog Turn 3}} & \multicolumn{3}{c}{{Dialog Turn 5}}  & \multicolumn{3}{c}{{Average}} \\
& \textbf{P}        & \textbf{R@10} & \textbf{R@50}     & \textbf{P}        & \textbf{R@10} & \textbf{R@50}     & \textbf{P}        & \textbf{R@10} & \textbf{R@50}  & \textbf{P}        & \textbf{R@10} & \textbf{R@50} \\
\midrule
\multicolumn{7}{l}{\textbf{Dresses}}\\
Retriever (R) + User (R)              & 89.45              & 6.25 & 20.26                        & 97.49             & 26.95 & 57.78                       & \textbf{98.56}             & 39.12 & 72.21 & 95.17 & 24.11 & 50.08\\
Retriever (R) + User (T)    & 89.10            & 7.00 & 21.28                       & 97.16             & 29.07 & 59.16                       & 98.18             & 41.57 & 70.93 & 94.81 & 25.88 & \textbf{59.46}\\
Retriever (T) + User (R)             & 92.29              & 11.61 & 33.92                       & \textbf{98.12}             & 36.18 & \textbf{69.34}                       & 98.52            & \textbf{42.40} & \textbf{74.78} & 96.31 & 30.06 & 59.35\\
Retriever (T) + User (T)             & \textbf{93.14}              & \textbf{12.45} & \textbf{35.21}                       & 97.96             & \textbf{36.48} & 68.13                       & 98.39             & 41.35 & 73.63 & \textbf{96.50} & \textbf{30.09} & 58.99\\

\midrule
\multicolumn{7}{l}{\textbf{Shirts}}\\
Retriever (R) + User (R)               & 89.39              & 3.86 & 13.95                        & 97.40             & 21.78  & 47.92                       & \textbf{98.48}             & 32.94 & 62.03 & 95.09 & 19.53 & 41.3 \\
Retriever (R) + User (T)             & 90.45            & 4.77 & 16.45                       & 97.14             & 20.52 & 46.60                       & 98.15             & 30.12 & 58.85 & 95.25 & 18.47 & 40.63\\
Retriever (T) + User (R)             & 91.77            & 9.33 & 27.15                       & 98.02             & 27.25 & 57.68                       & 98.41             & 30.79 & 62.53 & 96.07 & 22.46 & 49.12\\
Retriever (T) + User (T)             & \textbf{92.75}              & \textbf{11.05} & \textbf{28.99}              & \textbf{98.03}             & \textbf{30.34} & \textbf{60.32}              & 98.28             & \textbf{33.91} & \textbf{63.42} & \textbf{96.35} & \textbf{25.10} & \textbf{50.91}\\

\midrule
\multicolumn{7}{l}{\textbf{Tops\&Tees}}\\
Retriever (R) + User (R)                & 87.89             & 3.03 & 12.34                         & 96.82             & 17.30 & 42.87                       & 98.30             & 29.59 & 60.82 & 94.34 & 16.64 & 38.68 \\
Retriever (R) + User (T)             & 90.31              & 5.75 & 18.10                       & 97.73             & 27.72 & 56.42                       & \textbf{98.33}             & \textbf{36.20} & \textbf{65.45} & 95.46 & 23.22 & 46.66\\
Retriever (T) + User (R)             & 92.24              & 10.67 & 29.97                       & \textbf{97.90}             & 29.54 & 58.86                       & 98.26             & 33.50 & 63.49 & 96.13 & 24.57 & 50.77\\
Retriever (T) + User (T)             &  \textbf{93.03}            & \textbf{11.24} & \textbf{30.45}                        & 97.88             & \textbf{30.22} & \textbf{59.95}                       & 98.22             & 33.52 & 63.85 & \textbf{96.38} & \textbf{24.99} & \textbf{51.42} \\

\bottomrule
\end{tabular}
}
\end{center}
\caption{
    \textbf{Dialog-based Image Retrieval.} We report the performance on ranking percentile (P) and recall at N (R@N) at the $1$st, $3$rd and $5$th dialog turns. R / T indicate RNN-based and Transformer-based models.
    \label{table:additional_retrieval}
    }
\vspace{-1em}
\end{table*}

\end{document}